%% file: main.tex
\newif\ifarxiv
\arxivtrue
\documentclass{article}

\usepackage{microtype}
\usepackage{graphicx}
\usepackage{wrapfig}
\usepackage{booktabs} 

\usepackage[hyphens]{url}
\usepackage[breaklinks,bookmarks=false]{hyperref}

\usepackage[dvipsnames]{xcolor}

\usepackage[accepted]{icml2024}

\usepackage{amsmath}
\usepackage{amssymb}
\usepackage{mathtools}
\usepackage{amsthm}
\usepackage{xspace}
\usepackage[none]{hyphenat}

\usepackage[utf8]{inputenc} 
\usepackage[T1]{fontenc}    
\usepackage{amsfonts}       
\usepackage{nicefrac}       
\usepackage{subfig}
\usepackage{multirow}
\usepackage{balance}

\usepackage{enumerate}
\usepackage{tabularx}
\usepackage{nicefrac}
\usepackage{times}
\usepackage{enumitem}
\usepackage{caption}
\usepackage{subcaption}

\theoremstyle{plain}

\theoremstyle{definition}

\theoremstyle{remark}

\DeclareMathOperator{\Throughput}{Throughput}

\newcommand{\cA}{\mathcal{A}}

\newcommand{\phaze}{\textsc{Phaze}\xspace}

\newenvironment{packed_itemize}{
\begin{list}{\labelitemi}{\leftmargin=1em}
\vspace{-5pt}
 \setlength{\itemsep}{0pt}
 \setlength{\parskip}{1pt}
 \setlength{\parsep}{0pt}
}{\end{list}}

\icmltitlerunning{Integrated Hardware Architecture and Device Placement Search}

\begin{document}
\onecolumn
\icmltitle{Integrated Hardware Architecture and Device Placement Search}




\vspace{-2ex}

\begin{icmlauthorlist}
\icmlauthor{Irene Wang}{gatech}
\icmlauthor{Jakub Tarnawski}{msr}
\icmlauthor{Amar Phanishayee}{msr}
\icmlauthor{Divya Mahajan}{gatech}

\end{icmlauthorlist}

\vspace{2ex}


\icmlaffiliation{msr}{Microsoft Research, WA, USA}
\icmlaffiliation{gatech}{Georgia Institute of Technology, GA, USA}

\icmlcorrespondingauthor{Irene Wang}{irene.wang@gatech.edu}

\icmlkeywords{Machine Learning, ICML}

\vskip 0.1in




\printAffiliationsAndNotice{}  




%

\newcommand{\niparagraph}[1]{\vspace{0pt}\noindent\textbf{#1}}




\input{body/abstract}
\input{body/intro}
\input{body/related}
\input{body/architecture}
\input{body/overview}
\input{body/ilp}
\input{body/piperplus}
\input{body/results}
\input{body/conclusion}
\input{body/impact}
\input{body/acknowledgements}

\footnotesize




\balance
\bibliographystyle{icml2024}
\bibliography{cite}


\newpage

\normalsize
\appendix
\onecolumn
\input{body/appendix}


\end{document}

%% file: body/abstract.tex
\begin{abstract}

Distributed execution of deep learning training involves a dynamic interplay between hardware accelerator architecture and device placement strategy. 
This is the first work to explore the co-optimization of determining the optimal architecture and device placement strategy through novel algorithms, improving the balance of computational resources, memory usage, and data distribution.
%
%
%
Our architecture search leverages tensor and vector units, determining their quantity and dimensionality, and on-chip and off-chip memory configurations. 
It also determines the microbatch size and decides whether to recompute or stash activations, balancing the memory footprint of training and storage size. 
For each explored architecture configuration, we use an Integer Linear Program (ILP) to find the optimal schedule for executing operators on the accelerator.
The ILP results then integrate with a dynamic programming solution to identify the most effective device placement strategy, combining data, pipeline, and tensor model parallelism across multiple accelerators.
%
%
%
Our approach achieves higher throughput on large language models compared to the state-of-the-art TPUv4 and the Spotlight accelerator search framework. The entire source code of \phaze is available at {\small{\href{https://github.com/msr-fiddle/phaze}{\texttt{https://github.com/msr-fiddle/phaze}}}}.

\end{abstract}

%% file: body/intro.tex
\section{Introduction}

Deep learning training, due to its unique data flow, memory, and compute requirements of modern models, is often executed on specialized hardware known as domain-specific accelerators~\cite{tpuv4_isca, tabla, cosmic:micro}.
Moreover, due to the large memory footprint, training also needs to be executed in a distributed manner. 
Distributed deep learning divides a workload along various dimensions such as data parallel, pipeline parallel, and tensor model parallel, which mitigates the large memory requirements of training while boosting the throughput.
When designing a hardware accelerator for a single or set of models, an important question arises: "what architecture and model distribution strategy can achieve the optimal performance for end-to-end deep learning training?"

However, these two crucial aspects -- the model distribution strategy and the specific hardware -- have generally been examined in isolation. 
Some studies focus on identifying the most suitable architecture for deep learning execution, typically only for inference tasks ~\cite{confuciux, fast, spotlight}.
Others propose strategies for distributing models across accelerators, assuming a fixed domain-specific architecture~\cite{flexflow, pip}. 
The hardware architecture search explores the on- and off-chip resource utilization, whereas device placement strategy search offers a balance between memory footprint, networking overhead, and overall training throughput.
Thus, only performing architecture search~\cite{wham, dana} with a fixed device placement strategy can lead to under-utilization of the accelerator. 
Whereas, performing a device placement strategy on a fixed architecture~\cite{piper} might only search through a sub-optimal space of memory footprint and networking overhead. 
%
%
\emph{As such, identifying an optimal solution that co-optimizes accelerator architecture and distributed execution strategy for deep learning, remains a significant, yet unresolved, research challenge.}

To tackle this challenge, we introduce \phaze, a novel framework for co-optimizing hardware architecture, device placement strategy, and per-chip operator scheduling. 
\emph{This combined exploration of architecture configurations and its  schedule is complex, particularly when execution is distributed across multiple devices, resulting in a computationally vast multi-dimensional search space.}
Thus, to determine an accelerator architecture, \phaze~utilizes a hardware template derived from previous works, including both tensor and vector cores~\cite{tpu, tandem:asplos:2024}.
The hardware template defines the scope of the architecture search, with tensor cores handling matrix multiplication operators and vector cores executing point-wise and activation functions. 
For the device placement search, \phaze~determines model splits using a combination of pipeline parallel~\cite{pipedream}, intra-layer parallel or tensor model parallel~\cite{shoeybi2020megatronlm}, and data parallel~\cite{dataparallel} techniques.

Based on the hardware template, \phaze iterates through various architecture configurations. For each configuration, a novel Integer Linear Program (ILP) finds the optimal schedule for executing operators on a single accelerator, leveraging multiple cores when beneficial. 
\emph{Although ILPs generally produce optimal solutions, their inefficiency often limits their applicability. To avoid computational bottlenecks, our novel formulation circumvents the use of time-indexed variables, and its size depends only on the number of operators executed on an accelerator.}
%

The output of the ILP is then fed to a novel dynamic programming algorithm to determine the device placement, combining data, pipeline, and tensor model parallelism across multiple accelerators.
Additionally, \phaze~explores the trade-off between the memory required for training these models and the storage per accelerator by determining the microbatch that resides on each device and whether activations are recomputed or stashed~\cite{activationrecomputation}.

The exploration of all possible architectural configurations, even with a predefined hardware template, can be vast and computationally intensive. This process involves estimating latencies per operator, executing ILP to find the optimal latency and schedule of operators on an accelerator, and employing dynamic programming to determine the device placement strategy.
To address this, we introduce a heuristic that establishes an early stopping criterion based on the accelerator area and the performance metrics realized by the already explored configurations.

\emph{Overall, \phaze~is the first work to explore a large search space in executing distributed training, encompassing accelerator architecture, per-device scheduling, memory footprint and storage, and device placement.}
Our results show that the \phaze-generated architecture and device placement strategy for a set of large language models (Bert, OPT, Llama2, and GPT variants), on average, offer a $2.9\times$ higher throughput compared to a TPUv4 architecture with expert device placement.
Even when TPUv4 is augmented with the proposed device placement algorithm, the \phaze generated configuration offers $1.8\times$ higher throughput.

%% file: body/related.tex
\section{Background and Related Work}


\subsection{Distributed Training of Large Models}

The emergence of large models has necessitated various modes of parallelism that maintain the training fidelity while improving training throughput. 
Pipeline parallel training distributes layers across devices to reduce per-device storage requirements, processing mini-batches as micro-batches in a pipeline~\cite{pipedream-flush, gpipe}. 
As model sizes increase, there is a growing trend towards splitting a single layer across multiple accelerators, referred to as intra-layer tensor model parallelism.
%
%
%
%
For instance, Megatron-LM~\cite{shoeybi2020megatronlm} distributes a transformer layer by partitioning the self-attention and MLP layers over multiple devices. 
%
%
Data parallelism replicates the entire model multiple times, working alongside pipeline and tensor model parallelism. 
%
%
Additionally, certain runtime techniques, such as activation recomputation, minimize memory usage by recomputing activations during the backward pass instead of storing them, reducing memory needs but requiring the execution of the forward pass twice.
%
%

\niparagraph{Prior device placement techniques.}
Various works aim to determine the model partitioning strategy, utilizing diverse approaches such as RL-based~\cite{rl-deviceplacement}, MCMC-based~\cite{flexflow}, and dynamic programming-based~\cite{pip, piper} techniques to perform device placement.
FlexFlow~\cite{flexflow} enables data, tensor, and inter-layer parallelism; PipeDream~\cite{pipedream} facilitates data and pipeline parallelism; while Piper~\cite{piper} and Alpa~\cite{alpa} explore data, tensor, and pipeline parallelism.
Certain prior works do not search for the partitioning strategy, instead propose a model-specific fixed placement~\cite{fae, hotline}.
In \phaze, we not only propose novel algorithms that explore data, tensor, and pipeline parallelism for training through a dynamic programming algorithm, but also explore intra-operator parallelism on a single accelerator via an ILP formulation.
Moreover, previous studies assume a specific accelerator while determining the distribution strategy, and none optimize the accelerator architecture in conjunction with device placement.

\begin{figure*}
\centering
\includegraphics[width=1\textwidth]{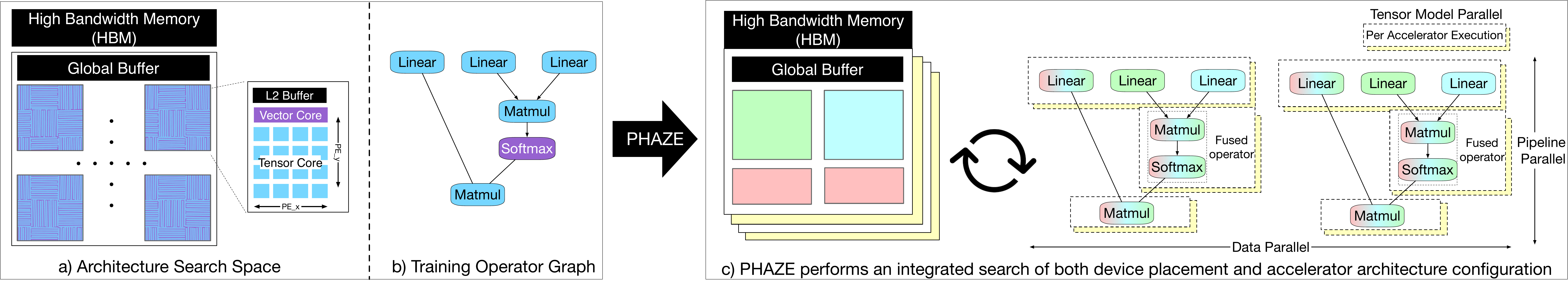}
\caption{(a) The template for an accelerator architecture consisting of hierarchical compute units, on-chip buffers, and off-chip HBM. A core can be of the type tensor, vector, or fused. (b) An example training operator graph that is used to optimize the accelerator and the distribution strategy. (c) The combined search space explored by \phaze. }
\label{fig:search_space}
\vspace{-3ex}
\end{figure*}

\subsection{Hardware Acceleration of Deep Learning}
Hardware accelerators are often used to execute deep learning due to their ability to handle predictable computation and memory access patterns~\cite{eyerissv2, transformer_Acc, nvdla}.
As such, accelerator vendors across the industry have largely converged on two types of cores to execute the operators relevant to these models: tensor and vector cores~\cite{tpuv4_isca, brainwave, tandem:asplos:2024}. Tensor cores handle high-throughput matrix operations, such as convolutions, General Matrix Multiplications (GEMMs), and batched matrix multiplications. Vector cores  perform point-wise and activation functions such as GELU, ReLU, and Tanh.
Each of these cores has access to memory buffers that feed and store activations, intermediate values, and parameters.



\niparagraph{Prior works on deep learning accelerator architecture search.}
Previous research in this field has developed architecture search frameworks to address the challenges of designing accelerators for continually evolving deep learning models. 
WHAM~\cite{wham} only performs architecture search for training deep learning models using a critical-path-based approach. 
It focuses solely on architecture search assuming a memory-based device placement strategy, hence, it does not optimize the model distribution. 
Additionally, WHAM’s ILP scheduling uses time-indexed variables, causing the ILP constraints to grow with the required training time, and results in significantly longer solving times. 
Whereas, \phaze’s ILP circumvents the use of time-indexed variables by encoding operation finishing time as a continuous variable, making the ILP more efficient.

Additionally, many prior works in this area exclusively address inference~\cite{dnnweaver, fast, prime, confuciux, spotlight, hasco}.
Fast~\cite{fast} utilizes Vizier~\cite{vizier}, a black-box optimizer, to generate hardware parameters and apply a linear program to solve their graph optimization problem. Prime~\cite{prime} employs a machine learning-based technique to reduce hardware simulations.
Both of these works focus on solutions like ILP formulation and accelerator design optimization primarily for forward operator execution, scheduling, and mapping, neglecting optimization for backward pass operators. However, in training, both forward and backward pass operators reside on the same device, necessitating optimization for both.
Other works ~\cite{confuciux, spotlight, hasco, apollo} focus solely on optimizing tensor cores for GEMM/CONV operators, overlooking the importance of non-linear operators like dropout and softmax in training.
Overall, \phaze is the first work to co-optimize hardware architecture and device placement strategy.
%


%% file: body/architecture.tex
\section{Integrated Hardware Architecture and Device Placement Search}

Deep learning acceleration raises a pivotal question: ``What hardware accelerator architecture best suits a particular deep learning model or a set of models, and how should these models be distributed across multiple devices?''
Figure~\ref{fig:search_space}(a) shows the accelerator template that outlines the scope of the architecture search.
This architectural template, along with the operator graph of the models (Figure~\ref{fig:search_space}(b)), is fed into \phaze~to determine the core configuration (quantity and dimensions), on-chip buffer and off-chip memory sizes and their bandwidths, the scheduling of each operator, and the distribution strategy of the entire training process. 
The search space explored by \phaze~is shown in Figure~\ref{fig:search_space}(c). 

\subsection{Search Parameters}

\niparagraph{Accelerator architecture search and runtime training parameters.}
The parameters and bounds defining the architecture search are illustrated in Table~\ref{tab:search_params}.

\begin{wraptable}{r}{9cm}
\centering
\caption{Architecture and training search parameters explored in \phaze~for per device execution. }
\includegraphics[width=0.5\columnwidth]
{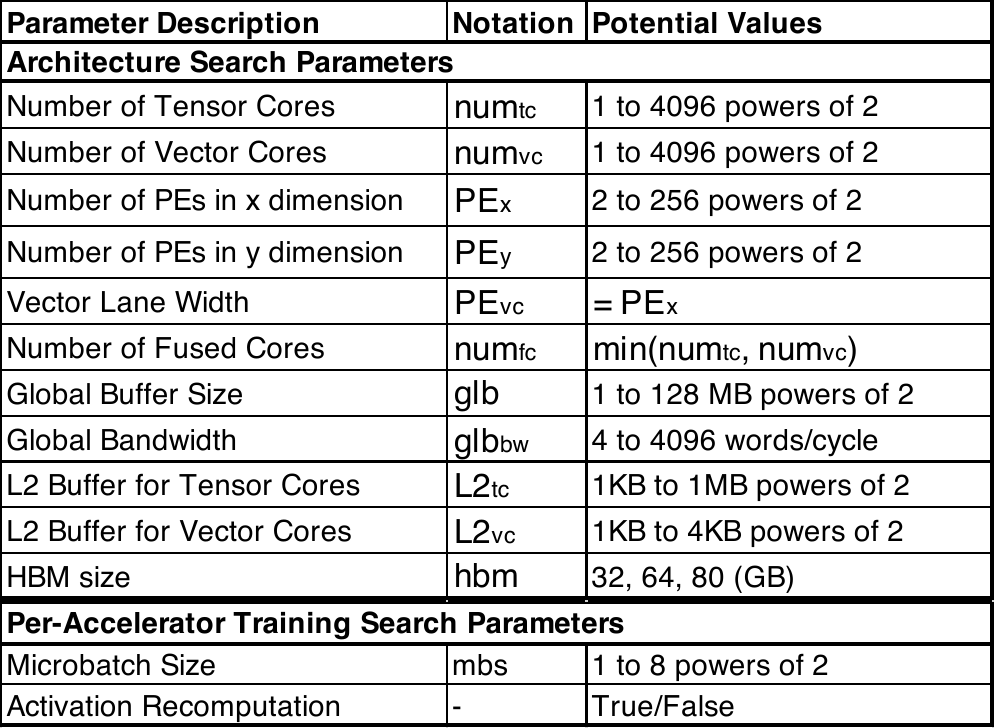}
\label{tab:search_params}
\end{wraptable}
\phaze's architecture search is defined by a template that aligns with industry-standard accelerators~\cite{tpuv4_isca, brainwave} and bounds that align with prior research~\cite{fast, prime}.
This template includes tensor cores for matrix multiplication and vector cores for activation functions, with their corresponding on-chip memory buffers for inputs and outputs. 
Similar to cloud-deployed accelerators and GPUs, the template features a closely coupled High Bandwidth Memory (HBM) for storing model parameters and activations~\cite{h100-nvidia, maia_microsoft}.
%
%
Each architecture configuration comprises two sets of parameters: the compute engine and memory configuration. 
The compute engine parameters, represented as a 5-tuple $\{num_{tc}, num_{vc}, PE_x, PE_y, PE_{vc}\}$, denote the number of tensor cores, number of vector cores, x- and y-dimensionality of the MAC units in each tensor core, and the width of the vector lane in each vector core. 
The on-chip memory configuration, $\{glb, glb_{bw}, L2_{tc}, L2_{vc}\}$, denote the global buffer size, the global buffer bandwidth, tensor core's L2 buffer size, and vector core's L2 buffer size. Additionally, $\{hbm\}$ represents the off-chip memory size.
To explore the trade-off between the memory footprint of training and the HBM size,  \phaze explores runtime training configurations, including microbatch size and training with activation recomputation or stashing.

%% file: body/overview.tex
\subsection{\phaze ~Workflow}


\niparagraph{Search problem definition.}
The global problem is to find the best hardware architecture configuration for an accelerator, where multiple accelerators execute distributed training.
The per accelerator search is bounded by an area constraint based on the configuration of the TPUv4 accelerator ~\cite{tpuv4_isca} with eight 128 $\times$ 128 tensor cores, two 128 wide vector cores, a global buffer of 128MB. The HBM is capped at 80GB.
%
%
\emph{The goal is to determine the optimal allocation of the area and HBM, focusing on the number of tensor cores, the number of vector cores, the width and depth of each tensor core, and the width of each vector core lane, to accommodate the training of a single or set of deep learning models}.
A model often has a combination of similar tensor dimensions throughout, determined by static hyperparameters such as attention heads, sequence length, hidden size, batch size, etc.
As a result, both activation and GEMM operators share similar dimensions. Hence, the vector and tensor core PE widths are restricted to be identical, i.e., $PE_x$ is equal to $PE_{vc}$.

\niparagraph{Outermost algorithmic problem.}
Out of all possible combinations in Table~\ref{tab:search_params}, we only consider the set of feasible architectures, i.e., those that meet the area constraint, and denote them as $\cA$.
%
%
In this work, to cater to training, we use throughput as the optimization objective.
The global objective, given oracle access to a function $\Throughput()$, is to maximize $\Throughput(W,A)$ over $A \in \cA$.
Here, $W$ is an input workload (DNN model). More generally, we can consider multiple models of interest, and take e.g.~a weighted average of the throughputs of $A$ on them.

\begin{figure*}
\centering
\includegraphics[width=0.9\textwidth]{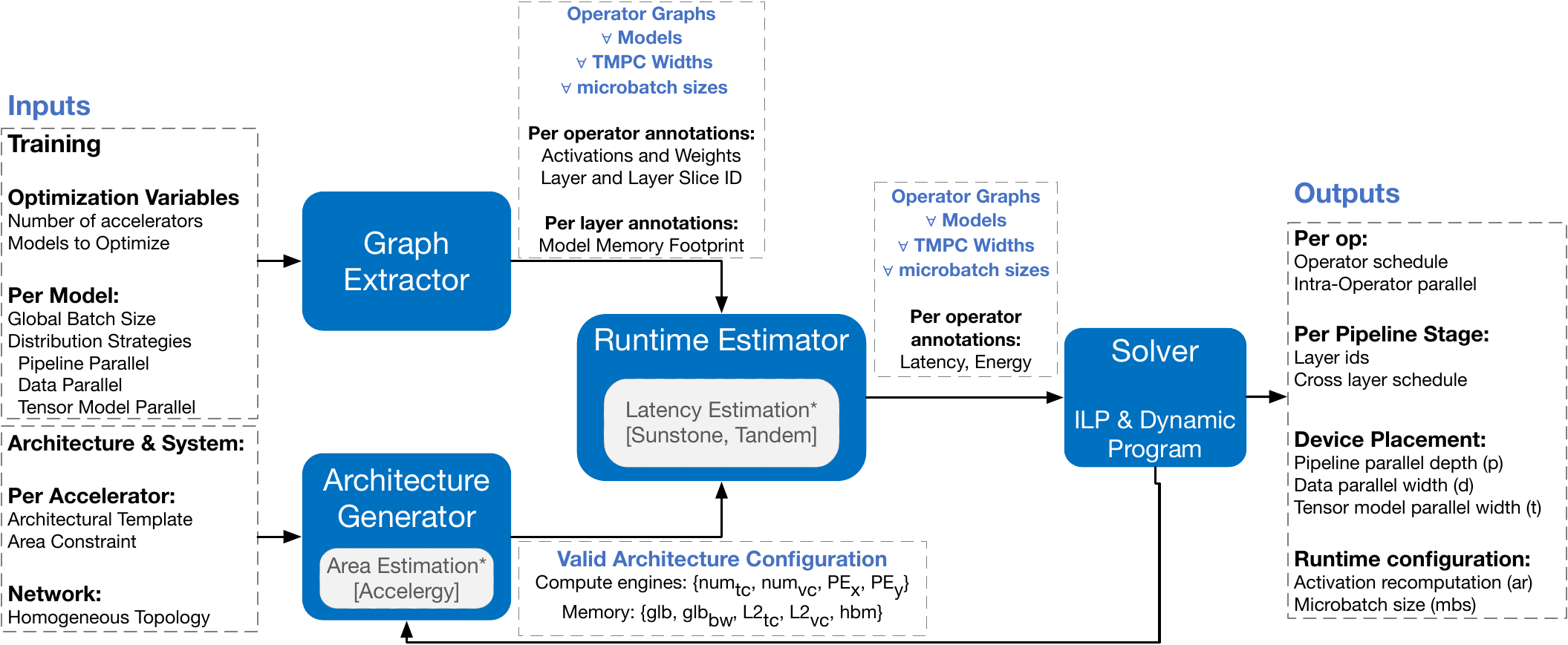}
\caption{The \phaze ~workflow -- the graph extractor extracts layer and corresponding operator graphs, which are annotated with memory footprint and latency estimates. The solver iteratively explores each valid architecture configuration.}
\label{fig:phaze}
\end{figure*}

\niparagraph{\phaze~Flow.}
Figure~\ref{fig:phaze} illustrates the \phaze workflow that solves the aforementioned optimization problem.

\niparagraph{Graph extraction.}
\phaze extracts the operator graph from model training scripts.
Among the data, pipeline, and tensor model modes of parallelism, the Tensor Model Parallel (TMP) width and the microbatch size impact the graph structure of the model. In TMP the tensor sizes are scaled according to the number of accelerators, and certain operators like AllReduce are added to combine interim results across accelerators. In this study, to evaluate \phaze we obtain graph slices based on the Megatron-LM~\cite{shoeybi2020megatronlm} strategy.
However, we stress that \phaze is a general framework that can work with any TMP technique (e.g.~MoE or sequence parallelism) as long as it is given the corresponding per-slice operator graphs as input.
Based on a mapping of operators to layers, the operator graph gives rise to a (much coarser) layer graph, as well as to an operator graph of each layer (or layer slice).
In this context, a layer refers to any continuous subgraph of the operator graph that is always placed on a single accelerator.

\ifarxiv
    In principle, the partitioning of an operator graph into layers can be arbitrary, as long as each layer constitutes a contiguous subgraph of the entire operator graph. Layers should not be chosen to be too large, as each layer is placed entirely on a single accelerator. On the one hand, a finer partitioning offers more flexibility in the pipeline model parallelism configurations found by the DP, as well as decreasing the ILP runtime (as layers are smaller). On the other hand, a coarser partitioning (larger layer subgraphs) enables the ILP to find more optimized operator schedules, while decreasing the DP runtime (as there are fewer layers).
\fi

\niparagraph{Architecture generator.}
The architecture generator provides the compute and memory tuple of each configuration explored  within the bounds of the architectural parameters provided in Table~\ref{tab:search_params} and the area constraint.
This architecture generator is devised as a heuristic based on accelerator configurations and their area. In this heuristic, we sort architectures in $\cA$ by area, starting with an initial accelerator that has the largest area.
The algorithm explores all architectures for $\Throughput()$ in the order of decreasing area. The exploration converges if the $\Throughput()$ trend diminishes with decreasing area of configurations. The best architecture out of those explored is then selected.
The details about the architectural exploration are provided in Appendix~\ref{apdx:arch_exploration}.

\niparagraph{Runtime estimator.}
The operator graphs (for all supported Tensor Model Parallel widths) are annotated with runtime estimates for each architecture being explored. 
We use existing libraries to determine the latency of the operator and the area of the architecture~\cite{sunstone, accelergy}.
For each operator, intermediate values are held in the Global and L2 buffers and the output data is transferred back to the HBM. At the start of each operator's execution, the required input activations and weights are transferred into the global buffer from the HBM.
These data transfer times are incorporated in the per-operator latency.

\niparagraph{Solver.}
Each layer (or layer slice) in forward and backward passes with its operator graph and runtime estimates is passed into the ILP solver to determine the optimal latency. This data, combined with the memory footprint of each layer (or layer slice), is consumed by the dynamic programming-based device placement to determine the optimal throughput.
The solver provides throughput feedback to the Architecture Generator to decide the next configuration to be explored, or to converge and identify the best accelerator configuration and corresponding device placement strategy. The algorithmic flow of \phaze is shown in Figure~\ref{fig:algorithm} in Appendix~\ref{apdx:phaze_algorithm}.
In the remainder of the paper (Sections~\ref{sec:ilp} and~\ref{sec:dp}), we focus on this solver, which computes $\Throughput(W,A)$ for a given workload $W$.

%% file: body/ilp.tex
\section{Integer Linear Program for Optimal Scheduling on a Single Accelerator}
 \label{sec:ilp}

\phaze's Integer Linear Program (ILP) determines the schedule that executes on a single accelerator. 
Accelerator execution involves a series of operators running on tensor or vector cores. 
Depending on the operator's compute intensity, it may execute on a single core or all cores, a scenario called intra-operator parallelism. 
For instance, a large GEMM is likely to benefit from utilizing all the available tensor cores.
In contrast, many independent but smaller operators might benefit from parallel execution (branching).
This balance is formulated as an ILP problem.
The ILP explores the search space of the schedule under the following condition: at any given moment, if intra-operator parallelism is utilized for an operator, then no other operators are in execution.
This assumption is made because the on-chip global buffer has a fixed bandwidth that can only feed the cores in a pipelined fashion. Therefore, if an operator is executing in intra-op parallel mode, the entire global buffer bandwidth is consumed by that operator.

\input{./body/figures/constraints}

The ILP is executed for every explored architecture configuration, and for every layer (layer slice) across all possible TMP widths and microbatch sizes of a model.
Thus, at this stage, we are not concerned with HBM usage, as the objective of the ILP is to schedule a layer or layer slice's operator graph on a single accelerator in order to minimize latency.
Instead, the dynamic program that determines the placement of layers (or layer slices) over multiple accelerators accounts for the HBM size, as explained in Section~\ref{sec:dp}.

\newcommand{\latencyy}[1]{\ell_{#1}}
\newcommand{\ioplatencyy}[1]{\hat{\ell}_{#1}}
\niparagraph{Input.}
The input is a Directed Acyclic Graph (DAG) $(V,E)$ of a layer (or a layer slice for tensor model parallel splits) where nodes $V$ correspond to operators, each with architecture-dependent latency estimates. 
\phaze incorporates an existing optimization technique called operator fusion that enables intermediate activations between certain operators to be directly forwarded from tensor to vector core, or vice versa, without passing through the HBM~\cite{tvm}.
Such an operator is executed on a fused core equipped with both a MAC unit and a vector lane.
As such, each operator in the graph is categorized as either tensor, vector, or fused. 
For each operator $i \in V$, we are given $\latencyy{i}$, its latency if not intra-operator parallelized, and $\ioplatencyy{i}$, its latency when intra-operator parallelized (i.e.~run on all tensor, vector, or fused cores).

Tensor cores are numbered from $1$ to $num_{tc}$, and vector cores from $1$ to $num_{vc}$. The first $\min(num_{tc}, num_{vc})$ tensor cores are paired with the first $\min(num_{tc}, num_{vc})$ vector cores. A fused operator, in the absence of intra-operator parallelism, runs on one such pair, i.e., on tensor core $c$ and on vector core $c$, for some $c \in \{1,...,\min(num_{tc}, num_{vc})\}$.

The edges $E$ signify data transfer to and from the HBM, which is accounted for in the operator estimates.

\niparagraph{Variables.} The ILP variables are enumerated below:
\vspace{-1ex}
\newcommand{\ztcc}{\ensuremath{z^{\mathrm{tc}}}}
\newcommand{\zvcc}{\ensuremath{z^{\mathrm{vc}}}}
\begin{packed_itemize}
    \item $t_i \in \mathbb{R}_+$ for $i \in V$: start time of operator $i$,
    \item $p_i \in \mathbb{R}_+$ for $i \in V$: latency of operator $i$ (defined by constraint \eqref{con:def_of_p_i}),
    \item $T \in \mathbb{R}_+$: the makespan (overall latency) of the schedule (defined by constraint \eqref{con:def_of_T}),
    \item $y_i \in \{0,1\}$ for $i \in V$: one if operator $i$ is intra-operator parallelized, zero otherwise,
    \item $x_{ij} \in \{0,1\}$ for $i, j \in V$, $i \ne j$: if this is one, $i$ finishes before $j$ begins,
    \item $\ztcc_{ic} \in \{0,1\}$ for $i \in V$ and $c = 1, ..., num_{tc}$: 1 if operator $i$ is assigned to tensor core $c$,
    \item $\zvcc_{ic} \in \{0,1\}$ for $i \in V$ and $c = 1, ..., num_{vc}$: 1 if operator $i$ is assigned to vector core $c$.
    \ifarxiv
        Here, intra-operator-parallelized operators are not thought to be assigned to any core; and fused operators are assigned to two cores. See constraints \eqref{con:z_assign_once}--\eqref{con:z_paired_cores}.
    \fi
\end{packed_itemize}
\vspace{-1ex}

\niparagraph{Constraints.}
We define the strict partial order $\prec$ as the transitive closure of the DAG $(V,E)$, i.e., we have $i \prec j$ if there is a path from $i$ to $j$ and $i \ne j$. If $i \prec j$, operator $i$ must finish before $j$ can begin, and only incomparable operators (those where neither $i \prec j$ nor $j \prec i$) can potentially execute simultaneously.

The novel idea of our ILP is that the variables $x_{ij}$ define another strict partial order, which lies between $\prec$ and the partial order given by the execution time intervals of operators in the found schedule. Specifically:

\begin{packed_itemize}
    \item if $i \prec j$, then $x_{ij} = 1$ (constraint \eqref{con:x1}),
    \item if $x_{ij} = 1$, then $i$ finishes before $j$ starts (constraints \eqref{con:prec}--\eqref{con:H}).
\end{packed_itemize}
\vspace{-2ex}

If $x_{ij} + x_{ji} = 1$, then $i$ and $j$ cannot execute in parallel. The $x$ variables enforce the condition that when intra-operator parallelized operators are executing, other operators cannot execute. This is done in \eqref{con:iop}: if $y_i = 1$ (i.e., $i$ is intra-operator parallelized), then we must have $x_{ij} = 1$ or $x_{ji} = 1$, which implies via \eqref{con:H} that $i$ and $j$ cannot execute in parallel.

In constraint \eqref{con:H}, $H$ is a large number. Intuitively, this constraint should be read as: $(t_i + p_i) \cdot x_{ij} \le t_j$ (which would be a quadratic constraint). Note that when $x_{ij} = 1$, \eqref{con:H} becomes $t_i + p_i \le t_j$, and when $x_{ij} = 0$, it becomes vacuous since the left-hand side is negative.

\newcommand{\ilpexplanations}{
    \item Constraints \eqref{con:antisymmetry} and \eqref{con:transitive} enforce that $x$ is a strict partial order.
    \item Constraint \eqref{con:x0} is implied by \eqref{con:x1} and \eqref{con:antisymmetry}; we keep it for clarity.
    \item In constraint~\eqref{con:H}, we set $H$ as the sum of all the node latencies, so that if $x_{ij} = 0$, the constraint is vacuous for any "reasonable" settings of $t_i$, $p_i$ and $t_j$. At the same time, $H$ should be kept from being too large to avoid numerical issues.
    \item Constraints \eqref{con:z_assign_once}--\eqref{con:z_paired_cores} ensure that $z$ is an assignment of non-intra-operator-parallelized operators to one core (or two paired cores in case of fused operators).
    \item Constraints \eqref{con:same_core_no_branch_TC}--\eqref{con:same_core_no_branch_VC} relate the $z$-variables to the rest of the ILP. They ensure that if two operators are assigned to the same core ($\ztcc_{ic} = \ztcc_{jc} = 1$ or $\zvcc_{ic} = \zvcc_{jc} = 1$), then one must execute before the other.
    \item For fused operators $i$, we could include the (implied) constraint $z_{ic} = 0$ for all cores $c$ that are "unpaired" (such cores exist if $num_{TC} \ne num_{VC}$).
}

\ifarxiv
    This main novel idea, together with the assumption described in the first paragraph of Section~\ref{sec:ilp}, allows us to encode operator finishing times as continuous variables (rather than indexing variables with time), which makes the ILP much more efficient than many previous attempts in the literature.
    
    We provide further explanations below:
    \begin{packed_itemize}
        \ilpexplanations
    \end{packed_itemize}

    Since clearly no more than $|V|$ cores can be utilized at any time for branching (without intra-operator parallelism), we need to use at most $O(|V|^2)$ $z$-variables, and thus the total number of variables in the ILP is $O(|V|^2)$.
    A further optimization that we employ is to include the $z$-variables (and the constraints involving them) only when necessary; see Appendix~\ref{apdx:ilp_constraint_reduction}.
\fi

%% file: body/figures/constraints.tex
\newcommand{\latency}[1]{\ell_{#1}}
\newcommand{\ioplatency}[1]{\hat{\ell}_{#1}}
    
\newcommand{\ztc}{\ensuremath{z^{\mathrm{tc}}}}
\newcommand{\zvc}{\ensuremath{z^{\mathrm{vc}}}}

 \begin{figure*}[h]
    {
    \begin{alignat}{4}
        \text{min}  \quad && T \nonumber \\
        \text{s.t.} \quad && T & \ge t_i + p_i & \quad & (\forall i) \label{con:def_of_T} \\
        && p_i & = y_i \cdot \ioplatency{i} + (1 - y_i) \cdot \latency{i} && (\forall i) \label{con:def_of_p_i} \\
        && x_{ij} + x_{ji} & \le 1 && (\forall i, j : i \ne j) \label{con:antisymmetry} \\
        && x_{ik} & \ge x_{ij} + x_{jk} - 1 && (\forall i,j,k : i \ne j, j \ne k, i \ne k) \label{con:transitive} \\
        && x_{ij} & = 1 && (\forall i \prec j) \label{con:x1} \\
        && x_{ji} & = 0 && (\forall i \prec j) \label{con:x0} \\
        && t_i + p_i & \le t_j && (\forall i \prec j) \label{con:prec} \\
        && t_i + p_i - H \cdot (1 - x_{ij}) & \le t_j && (\forall i, j \text{ incomparable}) \label{con:H} \\
        && x_{ij} + x_{ji} & \ge y_i &&  (\forall i, j \text{ incomparable}) \label{con:iop} \\
        && y_i + \sum_{c \le num_{tc}} \ztc_{ic} & = 1 && (\forall i : \text{ TC-operator or fused operator}) \label{con:z_assign_once} \\
        && y_i + \sum_{c \le num_{vc}} \zvc_{ic} & = 1 && (\forall i : \text{ VC-operator or fused operator}) \label{con:z_assign_once_VC} \\ 
        && \zvc_{ic} & = 0 && (\forall i : \text{ TC-operator}) (\forall c \le num_{vc}) \\
        && \ztc_{ic} & = 0 && (\forall i : \text{ VC-operator}) (\forall c \le num_{tc}) \\
        && \ztc_{i,c} & = \zvc_{i, c} && (\forall i : \text{ fused operator}) (\forall c \le \min(num_{tc}, num_{vc})) \label{con:z_paired_cores} \\
        && x_{ij} + x_{ji} & \ge \ztc_{ic} + \ztc_{jc} - 1 && (\forall i, j \text{ incomparable}) (\forall c \le num_{tc}) \label{con:same_core_no_branch_TC} \\
        && x_{ij} + x_{ji} & \ge \zvc_{ic} + \zvc_{jc} - 1 && (\forall i, j \text{ incomparable}) (\forall c \le num_{vc}) 
        \label{con:same_core_no_branch_VC}
    \end{alignat}
    \caption{ILP constraints. The optimization objective is to minimize the total latency/makespan $T$ of the layer (layer slice).}
    \label{fig:ilp}
    }%
\end{figure*}

%% file: body/piperplus.tex
\section{Device Placement to Maximize Throughput}
\label{sec:dp}

For a given accelerator architecture $A$, number of accelerators $K$, and a workload $W$, this stage of the solver aims to form a high-throughput pipelined schedule for the workload. 
For this problem, the workload is a Directed Acyclic Graph (DAG) $W=G=(V,E)$ with a \emph{layer} granularity (i.e., $V$ is the set of layers of the DNN) given for each supported TMP width $t$ and microbatch size $b$.
We assume the minibatch size is provided by the user:
$B$ is the number of microbatches in a minibatch. 
The algorithm is executed with activation recomputation both disabled (stashing) or enabled.
When enabled, it is used throughout the entire workload, with a stage granularity (that is, we store only the input activations of a stage, not of each layer, and compute the forward-plus-backward pass of the entire stage, materializing all the intermediate activations for a single microbatch during this time).
In this algorithmic problem the goal is to determine:
\begin{packed_itemize}
    \item the TMP width $t$ and microbatch size $b$,
    \item the data parallel width $d$ (number of parallel pipelines), 
    with $d \le \min(K,B)$,\ifarxiv\footnote{It might make sense to require $d \mid K$ and $d \mid B$, as otherwise some accelerators would be unused or microbatches would be split unevenly among the parallel pipelines.}\fi
    \item a number $s$ of stages, and a sequence of subgraphs/stages $S_1, ..., S_s$, each with a number of associated accelerators $K_1, ..., K_s$, such that:
        \begin{itemize}
            \item $S_1, ..., S_s$ form a disjoint partition of the set of layers: $V = S_1 \cup ... \cup S_s$,
            \item there are no edges from $S_i$ to $S_j$ with $i > j$,\ifarxiv\footnote{This in particular implies that each $S_i$ is a so-called contiguous subgraph.}\fi
            \item we have enough accelerators: $d \cdot \sum_{i=1}^s K_i \le K$,
        \end{itemize}
    \item for each stage $i=1,...,s$, a low-latency schedule to execute the layers in the subgraph $S_i$ using $K_i$ accelerators (that fits in accelerator memory).
\end{packed_itemize}

\niparagraph{Device placement algorithm.}
In the sequel we assume without loss of generality that the microbatch size $b$ is fixed; if multiple sizes are considered, we loop over them and select the best at the end.
The dynamic program builds the pipeline stage by stage, starting from the \emph{last} stage and ending with the first.
To do so, it works on \emph{downsets}: downward-closed sets of layers (i.e., a downset has no edge leaving it).
We build on the algorithm from Piper~\cite{piper}, with crucial extensions to handle flushing pipeline schedules, model the memory usage faithfully, and take advantage of branching in the layer graph.
In particular, we take flushing and gradient update synchronization into account using formulas~\eqref{deepak-formula} and~\eqref{dp-resync-cost} explained below.
Thus, 
we need to know the number of stages $s$ and number of data-parallel pipelines $d$.
However, our job will be made easier by the fact that both \eqref{deepak-formula} and \eqref{dp-resync-cost} are applied only when finishing building the pipeline in the dynamic program. To apply \eqref{dp-resync-cost}, we need to know the first stage, which our dynamic program allows for as we build stages from the last to the first.

We compare our device placement algorithm with Piper in Appendix~\ref{sec:comparison-piper} and with Alpa~\cite{alpa} in Appendix~\ref{sec:comparison-alpa}.

The dynamic programming table that we will compute is
$dp^t[D][k][s] := $ minimum max-load of any accelerator, when optimally partitioning downset $D$ over $s$ stages using $k$ accelerators, with TMP width $t$.
Note that at this level we are considering only a single (data-parallel) pipeline.
We will compute this for all downsets $D$ (except the entire set $V$) and numbers $k$, $s$ and $t$.
The \textbf{dynamic programming recursion} is as follows:

\[
    dp^t[D][k][s] = \min_{\text{downset } D' \subseteq D}
\min_a \max \left( dp^t[D'][k-a][s-1], load^t(D \setminus D',a,s) \right)
\vspace{-2ex}
\]
where:
\begin{packed_itemize}
    \item we are placing a new stage, with layer set $D \setminus D'$ and using $a$ accelerators,
    \item $load^t(S,a,s)$ is defined as the load (optimal, minimized latency) of a stage with layer set $S$, using $a$ accelerators, that is the $s$-th stage from the end of the pipeline, using TMP width $t$. An algorithm for computing $load^t(S,a,s)$ (where we also reason about the memory footprint of a stage) is provided in Section~\ref{sec:load},
    \item the remaining subproblem is to place the layer set $D'$ over $s-1$ stages using $k-a$ accelerators.
\end{packed_itemize}

We compute the final result by optimizing over $t$, $d$, $s$, and the first stage.
Namely, the \textbf{final time per batch, $F$}, is:

\footnotesize
\[
F :=
\min_t \min_d \min_s \min_{D'} \min_a \left[ (B/d + s - 1) \cdot  \max ( dp^t[D'][K/d-a][s-1],
load^t(V \setminus D', a, s)) + 4 \cdot \frac{d-1}{d} \cdot \frac{\text{weights}(V \setminus D')}{\text{bandwidth}} \right] 
\]

\normalsize

The expression for the final time per batch arises as follows:
\begin{packed_itemize}
    \item We form $d$ parallel pipelines. Thus we have $K/d$ accelerators to use per pipeline.
    \item We use $a$ accelerators (per pipeline) for the first stage, which has layer set $V \setminus D'$, where $D'$ is the downset. In an $s$-stage pipeline, it is the $s$-th stage from the end, and we loop over all $s$ values.
    \item The maximum load of any accelerator is given by the $\max$ expression.
    \item The other terms are responsible for pipeline flushes and gradient synchronization communication costs:
\end{packed_itemize}

\niparagraph{Pipeline flushes.}
The pipelining scheme we employ follows a \emph{flushing} schedule similar to PipeDream-Flush / DeepSpeed 1F1B~\cite{pipedream-flush}. This approach differs from previous work utilizing dynamic programming such as Piper, where non-flushing schedules like PipeDream-2BW were considered~\cite{piper}.
In a non-flushing schedule, the time taken per microbatch equals the maximum load (single-microbatch latency) of any stage. As such, previous research has concentrated on minimizing this max-load.

However, a major difficulty regarding flushing schedules is that the flush time (pipeline bubbles) cannot be entirely disregarded, especially if the minibatch size is small. We take into account the flush time using the following {approximation}: the time taken per batch is calculated by multiplying the max-load by a factor
%
\begin{align}
    \frac{B}{d} + s - 1. \label{deepak-formula} 
\end{align}

%
Note that $B/d$ is the per-pipeline batch size (number of microbatches per pipeline).
One can easily see that this approximation is lossless if all stages have the same load.

\niparagraph{Gradient synchronization communication costs.}
AllReduce communication, which synchronizes gradients between data-parallel replicas, takes place during the flush period. Throughout this time, all stages, except for the first, remain idle while the backward pass propagates. We {assume} this period is sufficient for the communication to successfully complete.
%
However, for the first stage, synchronization does cause a slowdown, which we account for by adding 
\footnotesize
\begin{align}
    4 \cdot \frac{d-1}{d} \cdot \frac {\text{size of weights in first stage}}{\text{bandwidth}}
\vspace{-2ex}
 \label{dp-resync-cost}
\end{align}
\normalsize

to the execution time of a batch.

The bandwidth here is the same as the one used to estimate an AllReduce operator cost for tensor model parallelism.

\subsection{Computing the Load for the Dynamic Programming Algorithm}
\label{sec:load}

It is important to note that, since the $load$ subroutine will be invoked for all feasible settings of $S,t,a,s$, it must be highly efficient. It computes the maximum load of any of the $a$ accelerators.
The second role of the $load$ subroutine is to calculate memory usage. If the memory limit (as defined by accelerator HBM size) is exceeded, $load$ should return $+\infty$.

Ideally, for every $s$, we would determine the schedule with the least latency such that no accelerator exceeds the memory limit. However, in general we do not solve this more complex problem. Instead, we attempt to identify the overall lowest-latency schedule, and calculate its memory usage.

\niparagraph{Compute and communication load.}

As in all of our test workloads the layer graph
(i.e., the operator graph of the full DNN, where operators belonging to each layer or layer slice have been contracted into a single node)
is linear (contains a Hamiltonian path -- this is consistent with the layer structure of large language models),
we only describe how to compute $load$ for this case.
We note that, as there are no optimization decisions to be made,
we in fact return an optimal layer latency.
We focus on the forward pass for simplicity; the computation for the backward pass (with or without forward pass recomputation) is analogous.

We fix a topological ordering of $S$ (recall that $S$ is a contiguous subgraph of layers).
Let $S = \{L_1, ..., L_\ell\}$ (in that topological order).
Each layer comes with a single, optimized way/schedule to execute it, which we compute as $schedule(L_i)$ in Section~\ref{sec:ilp}.
We will use an "object-oriented" notation to access the quantities related to this schedule, such as $schedule(L_i).latency\_fw$.
We denote similarly the quantities related to $L_i$ that are not dependent on the schedule, such as $L_i.weights\_size$.

We schedule the layers one by one. Each layer begins at the earliest time that all of its predecessors have been completed, and all incoming activations have been transferred over the edges of the graph from other devices (that is, we consider here the transfer costs on those edges that come from outside of the stage).
Our current assumption is that our network has a flat structure, where transmitting $X$ bytes from any accelerator to any other accelerator takes time $X/\text{bandwidth}$.
Then, the finishing time of the layer is the starting time plus $schedule(L_i).latency\_fw$.

\niparagraph{Calculating the memory footprint of training in pipelined execution.}

If activation recomputation is employed, it is performed at a stage level, meaning that all intermediate activations within the stage are materialized during the forward pass recomputation.
This implies that peak memory usage is reached at the end of that recomputation, specifically when the pipeline is in a steady state and the stage has stored data for the full number of in-flight microbatches.
At this point, the accelerator memory (HBM) contains the following:
\begin{packed_itemize}
    \item model weights,
    \item accumulated gradient updates (of the same size as the model weights),
    \item optimizer state,
    \item all intermediate activations,\footnote{Note that we do not differentiate between forward and backward activations, as they are of the same size. Moreover, as the backward pass progresses, the corresponding forward activations can be removed from memory, thus the memory usage will keep decreasing from the peak.}
    \item stashed data for in-flight microbatches; if activation recomputation is used, then these are the stage's input activations, otherwise, these are all intermediate activations.
\end{packed_itemize}


The following property of the implemented pipelining scheme determines the memory usage: in the steady state of the pipeline, for a stage that is the $s$-th stage from the pipeline's end, it's necessary to stash data for at most $s-1$ in-flight microbatches. These are the microbatches for which the forward pass has already been computed on this stage, but the backward pass has not yet been processed.
In the PipeDream-Flush scheme, computations can be  scheduled lazily, thereby satisfying the property. Here, the term "data" varies based on whether activation recomputation is being used: if it is not being used, the "data" refers to all forward activations, and if it is being used, "data" refers to the input activations of the stage.
For GPipe schedules, $s-1$ should be replaced with the total number of batches per pipeline, i.e., $B/d$. 

Therefore the memory usage can be modeled as:
\footnotesize
\[
\sum_{L_i \in S} \left( 2 \cdot L_i.weights\_size + L_i.optimizer\_size + L_i.activations\_size \right) + (s-1) \cdot stashed\_data
\]
\normalsize
where if activation recomputation is used, we have
\footnotesize
\[
stashed\_data = \sum_{(u,v) \in \delta^-(S)} size(u,v)
\]
\normalsize
where $\delta^-(S)$ denotes the set of incoming edges of $S$,
and otherwise
\footnotesize
\[
stashed\_data = \sum_{L_i \in S} L_i.activations\_size \,.
\]
\normalsize

\niparagraph{Dependence on $s$.}
Note that the computation costs and communication costs do not depend on $s$ (with the exception that the case $s=1$ is unique, as we do not require activation recomputation for the last stage). Moreover, the memory usage only depends on $s$ in an \emph{affine} way. Namely, when $s$ increases by one, the peak memory usage rises precisely by the amount of $stashed\_data$. This allows us to optimize the runtime of the $load$ computation by reusing the results across all $s$ values.
Indeed, rather than explicitly computing the quantity $load^t(S,a,s)$ for all $t,S,a,s$, we can instead return a pair $load^t(S,a)$ that comprises:
\begin{packed_itemize}
    \item the usual output of $load$ (maximum latency over the $a$ accelerators),
    \item the maximum $s$ for which the found schedule fits in the memory of every accelerator.
\end{packed_itemize}

\niparagraph{Range of $a$-values.}
Recall that in the dynamic program,
$a$ is the number of accelerators that handle a stage $S$ (set of layers).
We note that it is usually not necessary to loop over all possible values of $a$ from $1$ to the maximal available number of accelerators,
as the set of reasonable values of $a$ is much smaller.
Recall that we are working under a fixed TMP width $t$.
If $S$ contains a layer that admits TMP
(i.e., we have a layer slice operator graph for it),
then we need $a \ge t$, and otherwise, $a \ge 1$ is enough.
Beyond this, higher values of $a$ are only useful if $S$ contains enough (layer-level) branching to utilize more accelerators.
Therefore, in our evaluation workloads, which are Transformer-based LLMs,
it is enough to consider a single $a$-value.

\subsection{Runtime Analysis of the Dynamic Program}

We analyze the running time in terms of $O$-notation.
For simplicity and to model a practical scenario, we make the following assumptions:
\begin{itemize}
    \item The number of considered TMP widths $t$ is $O(1)$ (i.e., bounded by a constant).
    \item The number of considered microbatch sizes (that we loop over) is $O(1)$.
    \item The layer graph of the workload is linear (it contains a Hamiltonian path; e.g., it is a Transformer-based LLM).
    \item The number of edges in this graph is $O(|V|)$.
\end{itemize}
Then, using the above observations about restricting the dependence on $s$ and $a$, we can analyze the runtime as follows (recall that $K$ is the number of accelerators):
\begin{itemize}
    \item Precomputing $load$ values: there are $O(|V|^2)$ possible stages (contiguous layer subgraphs $S$), and computing $load$ for one stage takes $O(|V|+|E|) = O(|V|)$ time.\footnote{In fact, this should be possible to improve to an amortized runtime of $O(1)$ by computing $load$ for all stages of the form $\{l\}$, $\{l,l+1\}$, $\{l,...,l+2\}$, $\{l,...,l+3\}$, $\{l,...,l+4\}$ and so on in one go.} In total, we get $O(|V|^3)$.
    \item The main dynamic programming loop loops over values $D$, $k$, $s$ and $D'$, resulting in a runtime of $O(|V| \cdot K \cdot \min(K,|V|) \cdot |V|)$ (as we have $s \le K$ and $s \le |V|$).
    \item Computing the final time per batch takes time $O\left(\sum_{d=1}^K \frac{K}{d} \cdot |V|\right)$, where the $\frac{K}{d}$ term arises as we must have $s \cdot d \le K$, and evaluating the $\max$ expression takes $O(1)$ time (as we can precompute the $\mathrm{weight}(V \setminus D')$ terms in time $O(|V|^2)$). This gives $O(|V| K \log K)$ in total, which is dominated by the previous terms (as long as $\log K \le O(|V|)$).
\end{itemize}
In total, the runtime is $O(|V|^3 + |V|^2 K \min(K,|V|))$ (per each explored architectural configuration).

We remark that we made little attempts to optimize the dynamic program runtime. This is because the overall convergence time is anyway dominated by the operator estimations, not the solver.
To optimize the dynamic program runtime, one clear avenue would be to have a multi-threaded implementation, as the dynamic program is embarrassingly parallel (for example, for a given downset $D$, one could pursue all sub-downsets $D'$ in parallel). This should obtain almost linear scaling across CPU cores.

%% file: body/results.tex
\section{Evaluation}


We evaluate \phaze, the architecture search and solver, on a diverse set of large language models deployed in distributed training environments.
We obtain OPT~\cite{opt}, Bertlarge~\cite{bert}, GPT2~\cite{gpt}, and Llama2-7B~\cite{llama2} from the Hugging Face library~\cite{huggingface} and TMP graphs and hyper-parameters from public source code of Megatron-LM~\cite{megatronlm, shoeybi2020megatronlm}.
All the operator graphs are extracted using the Torch.fx library~\cite{torchfx} with microbatch sizes of 1, 2, 4, and 8.
Table~\ref{tab:workloads} shows the details of the evaluated workloads. 

\niparagraph{Operator level estimates.}
%
%
%
We use well-established toolchain Sunstone~\cite{timeloop, sunstone} for tensor core latency, Tandem for vector core latency~\cite{tandem:asplos:2024}, and Accelergy to determine the area of the accelerator for 22nm technology node~\cite{accelergy}.
Additional details are available in Appendix~\ref{apdx:op_estimates}.

\niparagraph{\phaze~execution.}
\phaze~is optimized over 1024 accelerators and a global batch size of 4096.
\phaze is executed on a V100 GPU and a Dual AMD Epyc 7713 CPU at 2.0 GHz with 128 cores, running Ubuntu 20.04.
The GPU runs CUDA 12.1 and is only used to extract the operator graphs. 
The overall \phaze~process is executed using Python 3.8.
The ILP formulations are solved using Gurobi 10.0.1~\cite{gurobi}.
The dynamic programming algorithm is implemented in C++, compiled with g++ version 11.3.0 and -O3 optimization flag.

\niparagraph{Baselines.}
We compare \phaze's architecture and device placement search with: (1) the TPUv4 architecture~\cite{tpuv4_isca}, which is the most commonly deployed accelerator for training, and (2) Spotlight, a state-of-the-art architecture search framework that uses Bayesian optimization~\cite{spotlight}. 
For meaningful comparisons, Spotlight has the same area constraint as TPUv4 architecture and uses the same toolchains, Accelergy and Sunstone for area and runtime estimations, respectively.

We assess the efficacy of \phaze's architecture search by comparing its performance against a fixed accelerator TPUv4 and Spotlight-generated designs, both executing with an expert placement strategy based on prior works~\cite{opt,efficient_Megatron}; more details in Appendix~\ref{apdx:results} Table~\ref{tab:workloads_apdx}.
To highlight the importance of co-optimizing both the architecture and the device placement, we compare \phaze~against TPUv4 and Spotlight-generated designs when they leverage \phaze's solver, the ILP layer scheduler, and the dynamic programming device placement.
For \phaze~and the baselines, we use a flushing schedule similar to PipeDream-Flush~\cite{pipedream-flush}.
%

\input{body/tables/workloads}


\begin{figure*}
\centering
\includegraphics[width=1\textwidth]{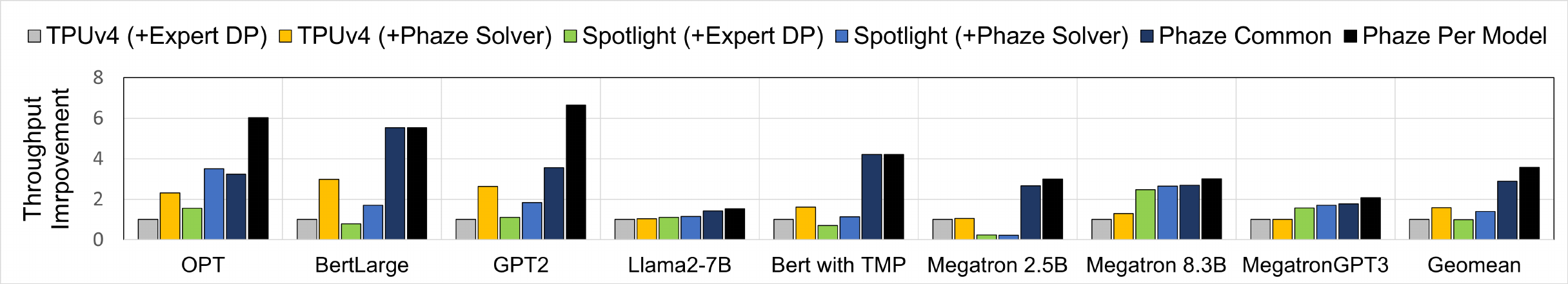}
\caption{Throughput comparison between the \phaze~Common and Per Model configuration with TPUv4 and Spotlight generated architectures. DP here is Device Placement.}
\label{fig:throughput}
\vspace{-3ex}
\end{figure*}

\subsection{Experimental Results}

Table~\ref{tab:workloads} shows the workloads evaluated by \phaze.
We use \phaze to generate a per-workload architecture and a common one across all the workloads. 
The \phaze-common architecture achieves high throughput across all models, with a compute configuration of $\{2, 512, 256, 256, 256\}$, on-chip memory configuration of $\{32, 4096, 1, 4 KB\}$, and an HBM of 64GB. We also present the geometric mean of all the throughput speedup results for each strategy compared to the TPUv4 baseline. This comparison assesses the overall performance efficiency of \phaze-searched systems across multiple models. Below, we explain the main observations and takeaways from the \phaze-Common and \phaze-per model architectures: 

%

\niparagraph{Area Utilization.}
The \phaze architectures are within 91\% of the area constraint, suggesting that for throughput, models gain from utilizing the majority of the area.
We observe that \phaze however does not require an area as large as TPUv4, albeit providing higher throughput.
Figure~\ref{fig:bert_hysteresis} in Appendix~\ref{apdx:phaze_search} provides further details.

%
%
%

%

\niparagraph{Compute Configuration.}
We observed that \phaze tends to favor larger tensor cores, typically with dimensions of 256, to accommodate GEMM operators in models. Phaze's common configuration has the same effective tensor core FLOPS as the TPUv4 configuration. However, larger tensor cores offer increased reuse in large models like GPT-3 compared to smaller cores, which require more coordination and local buffer memory optimizations for similar reuse. This underscores the advantage of larger tensor cores over a greater number of smaller ones.

\phaze also tends to choose architectures with a higher number of vector cores within the given area constraint. This facilitates the parallelization of operations like Layer Normalization, enhancing overall throughput. As a tradeoff, these selected architectures typically feature a smaller Global Buffer memory size (ranging from 4 to 32 MB) compared to the 128 MB of TPUv4.

%
%
%
%
%
\niparagraph{Global Buffer Bandwidth.}
Spotlight-searched configurations are limited to a single core for the tensor unit, hence the selected architectures lean towards a larger width. Despite this, these configurations have a significantly smaller global buffer bandwidth compared to \phaze architectures. This suggests that larger core sizes may not always be advantageous without sufficient global buffer bandwidth, resulting in lower throughput for Spotlight configurations despite wider cores.

\niparagraph{Memory configuration.}
We observe that \phaze~selected architectures do not select a $glb$ size greater than 32MB, indicating that the memory hierarchy with an L2 buffer can keep the cores utilized. 

\phaze selects the optimal HBM, choosing the smallest HBM necessary to achieve high throughput. Among the 8 workloads, only $Megatron~2.5B$ requires an 80GB HBM, to perform stashing rather than recomputation. 
%
%
For the remainder of the models, the throughput for 64GB and 80GB HBM configurations is either identical or provides only a marginal increase in throughput. 
Hence the \phaze-Common does not have the largest HBM. This indicates that larger memory does not always translate to higher throughput. Additionally, optimizations such as activation recomputation can mitigate the memory footprint of training these models.
%

\niparagraph{Throughput improvement.}
The \phaze-Per Model and \phaze-Common architectures and device placement strategies, on average, deliver a $3.6\times$ and $2.9\times$ higher throughput compared to TPUv4 with expert device placement strategy, respectively. 
We observe that both TPUv4 and Spotlight searched architectures achieve higher throughput when utilizing \phaze's device placement solver. This indicates that \phaze's algorithm further enhances the throughput of each model. However, this approach does not actually fully leverage the co-optimization feedback loop because the architecture is already fixed. In contrast, during \phaze's architecture search, the ILP-Dynamic Program solver guides the Architecture Generator by providing throughput feedback for exploring the next configuration. The \phaze-Common architecture and device placement strategy deliver $1.8\times$  and $2\times$  higher average throughput against the two baselines, respectively. This demonstrates that the co-optimization strategy of \phaze's architecture search and device placement algorithm enables a more optimized configuration that offers higher throughput for distributed training systems.

%
%
%
%

%

%

%

\niparagraph{Convergence time.}
The \phaze framework executes the following modules for each architecture: operator estimates, ILP per layer or layer slice, and the dynamic programming algorithm to optimize the model partitioning scheme.
The ILP, despite optimizing an operator scheduling, circumvents the use of time-indexed variables; it uses $O(|V|^2)$ variables, where $|V|$ is the number of nodes in the operator graph of a single layer or layer slice.
%
%
The largest model, Megatron GPT3, has $\sim$100 nodes in each layer and spends under 2\% of the execution time performing all the ILP optimizations. 
The dynamic programming optimization dominates the solving time as it is repeated for all HBM and recomputation/stashing configurations. 
Most of the execution time is spent on estimating operator latencies using the external library.
As such, models with TMP demand longer convergence times due to the higher number of explorations required.
Appendix~\ref{apdx:results} details the breakdown of the convergence time for each model. 

%% file: body/tables/workloads.tex
\begin{scriptsize}
\newcommand\ExtraSep
{\dimexpr\cmidrulewidth+\aboverulesep+\belowrulesep\relax}

\newcolumntype{?}{!{\vrule width 2pt}}
\setlength\extrarowheight{3pt}

\begin{table*}
\centering
\caption{Workloads evaluated using \phaze. The details of the model, the architecture configuration generated per workload, and the corresponding distribution strategy. The hyper-parameters are number of layers (\#L), attention heads (\#AH), and Hidden size (H). The distribution strategy is pipeline parallel depth (p), data parallel width (d), and TMP width (t), represented as $\{p, d, t\}$. The compute configuration is $\{num_{tc}, num_{vc}, pe_x, pe_y, pe_{vc}\}$, and the on-chip memory configuration is $\{glb, glb_{bw}, L2_{tc}, L2_{vc}\}$. Unless otherwise specified, all memory configurations are in MB.}
\resizebox{1.03\columnwidth}{!}
{\begin{tabular}{ c | c | c | c | c | c | c | c|c|c|c|c|c}
 \hline 
 &  &  & \textbf{Hyper} &  & \multicolumn{1}{r}{\textbf{Spotlight}}  &  & \multicolumn{5}{c}{\textbf{\phaze}}  \\   \cline{6-13} 
\textbf{Model} & \textbf{Model} & \textbf{Sequence} & \textbf{parameters} & \textbf{\textbf{TMP}} & \textbf{\textbf{Accelerator}} & \textbf{Memory} & \textbf{Accelerator} & \textbf{Memory} & \textbf{Device} & \textbf{mbs} & \textbf{Recomputation} & \textbf{HBM}\\ 
 & \textbf{Parameters} & \textbf{Length} & (\#L, \#AH, H) & \textbf{\textbf{Widths}} & \textbf{Configuration} & \textbf{Configuration} & \textbf{Configuration} & \textbf{Configuration} & \textbf{Placement} & & \textbf{vs. Stashing} & \textbf{size}\\
 \hline
OPT & 350M & 2048 & 24, 16, - & - & \{1, 1, 1024, 16, 1024\} &  \{128, 251, 26, 26\} &  \{4, 256, 256, 256, 256\}&  \{4, 4096, 1, 4KB\} &  \{1,1024,1\} & 2 & Stashing & 64 GB\\ 
BertLarge & 350M & 512 & 24, 16, 1024 & - & \{1, 1, 64, 256, 64\} & \{128, 251, 20, 20\}  & \{2, 512, 256, 256, 256\} & \{32, 4096, 1, 4KB\}  & \{1, 1024, 1\} & 4 & Stashing & 32 GB\\ 
GPT2 & 1.5B & 1024 & 48, 25, 1600 & - & \{1 ,1, 64, 256, 64\} & \{128, 252, 12, 12\} & \{1, 1024, 256, 256, 256\} & \{8, 4096, 1, 4KB\} & \{1,1024,1\} & 1 & Stashing & 64 GB \\ 
Llama2-7B & 7B & 4096 & 32, 32, 4096 & - & \{1 ,1, 8192, 2, 8192\} & \{128, 135, 12, 12\} & \{4, 1024, 256, 64, 256\} & \{8, 4096, 1, 4KB\} & \{6,164,1\} & 1 & Recomputation & 64 GB \\ 
\hline
\multicolumn{6}{l}{\textbf{Tensor Model Parallel Models}} \\
\hline
Bert with TMP & 350M & 512 & 24, 16, 1024 & 1,2,4,8 & \{1, 1, 64, 256, 64\} &  \{128, 251, 20, 20\} & \{1, 1024, 256, 256, 256\} & \{8, 4096, 1, 4KB\} & \{1, 256, 4\} & 8 & Stashing & 32 GB\\
Megatron 2.5B & 2.5B & 1024 & 54, 20, 1920 & 1,2,4 & \{1, 1, 16, 1024, 16\} &  \{128, 251, 26, 26\} &  \{1, 1024, 256, 256, 256\} & \{8, 4096, 1, 4KB\}  & \{2, 256, 2\} & 2 & Stashing & 80 GB\\
Megatron 8.3B & 8.3B & 1024 & 72, 32, 3072 & 1,2,4,8 & \{1, 1, 4096, 4, 4096\} &  \{128, 220, 22, 22\} & \{1, 1024, 256, 256, 256\} & \{8, 4096, 1, 4KB\} & \{2, 128, 4\} & 2 & Recomputation & 64 GB \\
Megatron GPT3 & 175B & 2048 & 96, 96, 12288 & 4,8 & \{1, 1, 4096, 4, 4096\} &  \{128, 147, 18, 18\} &  \{1, 1024, 256, 256, 256\} & \{8, 4096, 1, 4KB\} & \{16, 16, 4\} & 1 & Recomputation & 64 GB  \\\hline 

\end{tabular}}

\label{tab:workloads}
\end{table*}
\end{scriptsize}

%% file: body/conclusion.tex
\section{Limitations}
\textbf{Network topology and bandwidth}. We assume a homogeneous network and do not consider hierarchical or multi-level network topologies in \phaze including collective operations across tensor model parallel and data parallel execution.
\ifarxiv
    However, it may be possible to adjust our dynamic program to handle hierarchical network topologies similarly as Piper (see Appendix F in~\cite{piper}).
\fi

\textbf{Overlapping compute and communication}. ~\phaze leverages tiling within an operator by using toolchains such as Sunstone, where an operator is split across the cores to ensure compute and communication can be overlapped. The ILP further formulates this intra-operator split as an optimization problem to determine the per-accelerator schedule. However, for cross-operator execution, unless operators have been fused, ~\phaze assumes that if a communication operator follows the GEMM operator, they are not overlapped.

\textbf{Hardware Architecture Search}. \phaze offers a structured framework for exploring hardware architectures and existing methods to distribute models across accelerators. It can integrate new Tensor Model Parallelism strategies across accelerators, such as sequence-parallel or Mixture of Expert, but does not devise novel TMP strategies. \phaze also adopts globally set degrees of parallelism across all 
\ifarxiv
    layers (and thus across all stages).
\fi

\section{Conclusions}
\phaze~offers algorithmic solutions to perform the co-optimization between accelerator architecture search and model partitioning for distributed training.
\phaze makes the multi-dimensional optimization space of architecture search and device placement tractable by reducing the number of accelerator architectures explored through area-based heuristics and employing a novel Integer Linear Program (ILP), the complexity of which is dependent only on the number of operators in a single layer.
Uniquely, our ILP scheduling optimization also explores the partitioning of operators across cores, known as intra-operator parallelism. 
%
%
Based on the optimal backward and forward pass latencies, \phaze~then leverages a novel dynamic programming approach to determine the device placement and model partitioning scheme.

%% file: body/impact.tex
\section*{Impact Statement}

The research presented in this paper will advance the design exploration of AI supercomputers by providing enhanced tools capable of navigating the trade-off between performance and efficiency. This improvement is expected to reduce both time-to-market and development costs.
The outcome of this work will be the development of frameworks and tools that integrate the exploration of hardware design and distributing workloads and device placement for deep learning thus significantly lowering the barriers to developing next-generation deep learning infrastructure.
This is because such frameworks will reduce the necessity of deploying large models and conducting resource-intensive explorations involving compute, memory, and energy considerations to determine the architectural configuration and device placement strategy.
To achieve these goals, this paper has fostered closer interactions between the architecture, machine learning, systems and networking, and theoretical computer science communities.

%% file: body/acknowledgements.tex
\section*{Acknowledgements}

We thank the anonymous reviewers for their insightful comments. This research was supported in part through computational resources provided by Partnership for an Advanced Computing Environment (PACE) at Georgia Tech, Institute for Data Engineering and Science (IDEaS) at Georgia Tech, Google Cloud, and Microsoft Azure. This work was partially supported by Gifts from Google and AMD. The views and conclusions contained herein are those of the authors. They should not be interpreted as representing the official policies or endorsements, either expressed or implied, of Georgia Tech or Microsoft Research.

%% file: body/appendix.tex
\section*{Appendix - Integrated Hardware Architecture and Device Placement Search}

\ifarxiv
    \section{Reducing the Complexity of the ILP}
    \label{apdx:ilp_constraint_reduction}
\fi
Due to the use of $x$-variables that encodes a strict partial order that is in-between the input partial order $\prec$ and the partial order induced by the computed solution, we manage to circumvent the necessity of using time-indexed variables.
Therefore our ILP is very tractable. Below we describe an additional optimization that we employ.
The proposed ILP has two sets of constraints:
\begin{packed_itemize}
    \item ensuring the order and dependencies in the operator graph,
    \item ensuring the resource constraint imposed by the accelerator configuration (not too many cores are used at the same time).
\end{packed_itemize}

The number of variables in the Integer Linear Program (ILP) depends on the number $|V|$ of nodes in the layer or layer slice, as well as the number of cores in the accelerator (both tensor and vector cores). Namely, it is $O(|V|^2 + |V|(num_{TC} + num_{VC}))$.
The $z$-variables ensure that the breadth of parallel execution is not excessive -- that is, it does not employ more cores of either type than available.
However, it is often the case that this restriction is not necessary.
This is particularly likely if the number of tensor/vector cores is large.

We optimize the ILP as follows.
We first remove the $z$-variables and all constraints involving them, and solve the ILP.
We then build the schedule based on the ILP solution (start executing every operator $i$ at time $t_i$),
and check how many vector cores and tensor cores are used at any time.

If the found optimal solution does not use more cores of either type than are available, then we have an optimal solution.
Otherwise, it is necessary to add back the $z$-variables and constraints corresponding to the type (vector or tensor) of cores for which the resource constraint has been violated.
We then re-solve the ILP.

Note that whenever this happens, it must be the case that there is significant branching in the model (more nodes are executing in parallel than the number of tensor or vector cores).
This in particular implies that the number of cores is smaller than the size $|V|$ of the operator graph of the layer (or layer slice).
Therefore, when we proceed in this way, the number of variables in the ILP is always at most $O(|V|^2)$.
The ILP runtimes are also only a small fraction of the entire end-to-end compute time.

\section{Further Comparisons and Details on Dynamic Programming Based Device Placement} \label{sec:dp-details}

\subsection{Comparison to Piper's Algorithm} \label{sec:comparison-piper}
While our dynamic programming algorithm builds on Piper~\cite{piper}, it is not simply an extension or augmentation; rather, it takes a different direction.
While both build solutions stage-by-stage, our algorithm differs from Piper's in several key aspects:
\begin{itemize}
    \item Piper allows different tensor parallelism and data parallelism degrees at each stage, potentially leading to complex pipelining schedules. In contrast, we opt for globally set degrees of tensor parallelism and data parallelism, thus addressing realistic deployment scenarios.
    \item In the same vein, in Piper, even different layers in the same stage might use different TMP strategies (though of the same TMP degree) and different activation recomputation statuses (enabled / disabled). This leads to the need to heuristically solve an NP-hard knapsack subproblem as a subroutine in Piper's equivalent of the $load$ subroutine.
    \item \phaze's dynamic program facilitates practical pipelining schedules with flushes like 1F1B PipeDream-Flush or DeepSpeed by considering per-pipeline batch size and pipeline depth (see \eqref{deepak-formula}). This is made possible thanks to being cognizant of the per-pipeline batch size and the pipeline depth.
    \item \phaze's dynamic program supports branching across accelerators within a stage (if possible for a given workload; it is not possible for linear workloads such as Transformer-based LLMs).
    \item \phaze offers precise modeling of AllReduce costs related to data-parallel gradient update synchronization (see \eqref{dp-resync-cost}); Piper's modeling is overly pessimistic.
    \item Piper's runtime is $\tilde{O}(|V|^3 B + |V|^2 K B d(B))$, where $B$ is the number of microbatches in a minibatch, and $d(B)$ is its number of divisors. This is inferior to \phaze's dynamic program. Piper is not concerned with an architecture search, whereas we run our dynamic program for every considered architecture and thus require higher efficiency.
\end{itemize}

\vspace{-3ex}

\subsection{Comparison to Alpa's Algorithm}
\label{sec:comparison-alpa}

Alpa~\cite{alpa}'s approach is superficially similar in that it also uses an ILP inside a dynamic program.
However, Phaze’s ILP is unique in that it considers both branching and intra-op parallelism, and yet stays tractable despite this large search space. In the cost model used by Alpa, the latency of a layer (or sequence of consecutive layers) is defined as the sum of operator latencies plus communication costs, which precludes taking advantage of branching structure in the operator graph, or of overlapping communication with computation. Alpa also does not model the communication cost between stages in the DP, and does not support branching across accelerators in the DP. On the other hand, Alpa does attempt to automatically find certain tensor model parallelism strategies, whereas \phaze expects the TMP strategies on the across-accelerators level to be provided as input in the form of operator graph slices.

\section{\phaze~Integrated Search and Workflow}
\label{apdx:phaze_search}

In this section, we describe in detail the \phaze~workflow and solver, illustrated in Figure~\ref{fig:phaze}. 
Algorithm~\ref{alg:phaze} illustrates the process of generating architectures within the area constraint, extracting graphs for each model, estimating runtime latency, executing the \phaze~ILP and dynamic program solver, and converging on the architecture and the device placement algorithm through a feedback loop. We go into further detail for each step in the following subsections.

\input{body/algorithms/phaze}
\subsection{Architecture Generator}
\label{apdx:arch_exploration}

The search bounds for each architectures explored through \phaze~are listed in Table~\ref{tab:search_params}. 
However, there is an area constraint. The accelerator with the maximum possible area has the compute configuration of $\{8, 2, 128, 128, 128\}$, and a $glb$ of 128MB (TPUv4).
We only consider architecture within the area constraint and sort the architectures by area. 

\niparagraph{Generating architecture designs and area estimations}
To sort the configurations based on area, \phaze generates area estimates for all possible combinations of $\{num_{tc}, num_{vc}, PE_x, PE_y, glb\}$ in the search space. 
The $L2$ buffers, $L2_{tc}$ and $L2_{vc}$ are scaled based on the width and depth of the cores. 
This scaling is performed to meet the memory requirements of the tiling dataflow for generating a valid schedule and mapping for computations on the accelerator configuration.
The equation used for the scaling is: $L2 = 2^{log_2{PE_x} + log_2{PE_y} - 6}$ in KB. 
This yields a maximum buffer size of 1MB for Tensor cores when both $PE_x$ and $PE_y$ are at the maximum value, 256.
Because PEs in the vector core are arranged in a pipelined manner, the L2 buffer required to fully utilize the vector core is based on the width of the vector lanes. Given that the vector core's lanes are as wide as ${PE_x}$, the maximum L2 buffer size for a vector core is 4 KB.
If the scaling equation yields a value below 1~KB, the L2 buffer size is capped at a minimum of 1 KB. 
Additionally, $glb_{BW}$ has a maximum value of 4096 words per cycle and is scaled to ensure optimal utilization of all tensor cores.

\niparagraph{Pruning the number of accelerator configurations explored.}
Our architectural pruner is based on the following insight: utilizing the maximum area would allow us to either instantiate a larger number of cores or larger-sized cores.
As training is a throughput centric task with high memory footprint, it would intuitively benefit from more compute and memory.
Thus, we explore the architectures in a decreasing order of the area. 
A smaller area suggests a reduction in either the core size or the number of cores. If a smaller dimension does not yield a better training metric than the previous area configurations, it suggests that the tensor dimensions in the operators are large, and benefit from a higher number or larger dimensional cores. Alternatively, it could indicate that the parallelism in the operator graph is high, necessitating a greater number of cores.

To prevent settling for a local minimum, we introduce a hysteresis level in the pruner.
A hysteresis level determines the number of smaller areas that need to show a reduction in the metric (in this case, throughput) before the search process converges. The pseudo-code for this heuristic is provided in algorithm~\ref{alg:phaze} as the function $check\_converge$.
The tradeoff of the hysteresis level and the convergence time is shown in Appendix~\ref{apdx:hysteresis}.
A higher hysteresis level means more configurations are explored, however, with smaller areas. 

\subsection{Per-operator Estimates}
\label{apdx:op_estimates}

The architecture search in \phaze~relies on the per-operator estimates to determine the optimal schedule per layer or layer slice using the ILP.
A flexible and hardware-validated operator mapper is critical for evaluating accelerator performance for a given model.
As mentioned, each operator executes on a predefined core type. 
Table~\ref{tab:mapping} shows a small subset of common operators in deep learning and their core mapping. A tensor core operator followed by a vector core operator is fused and can only be executed on the fused core.

\niparagraph{Tensor, vector, and fused core operators.}
Operator estimates for tensor, vector, and fused core operations (excluding Allreduce) are generated using validated hardware tools like Sunstone, Tandem, and Accelergy~\cite{sunstone, tandem:asplos:2024, accelergy}. Sunstone and Tandem provide latency for tensor and vector core, respectively,  and Acceleragy estimates area.
For each operator and accelerator configuration, two estimates are produced:
\begin{packed_itemize}
\item The execution time when operated on a single core of the specified type
\item The execution time when operated on all cores of the type (termed "intra-operator latency")
\end{packed_itemize}
As Figure~\ref{fig:search_space}(a) illustrates, each core is an array of MAC units and/or vector lanes.
Intra-operator latency refers to the execution of a single operator across multiple cores. Tensor core or matrix multiplication based operators, in particular, often deal with multi-dimensional sizes.
%
\input{body/tables/op_mapping}

\niparagraph{Network and communication collective estimates.}
Collective operators are necessary in both tensor model parallel and data parallel execution. 
In data parallel execution, the model is replicated across devices, and weight updates are gathered across these pipelines.
The overhead of the collective operators due to data-parallel execution is accounted for by the dynamic programming algorithm as it optimizes for the distribution strategy.

Additionally, \phaze enables tensor model parallel execution, which splits a single layer across multiple accelerators. 
The accelerators participating in this form of parallelism need to combine results across the devices.
Thus, collective operators are introduced across layer slices in the tensor model parallel mode. 
To estimate latency for these operators, we assume that the network is homogeneous with a certain bandwidth and the collective is modeled in the following manner: 

 \footnotesize
\begin{align}
\nonumber 
    \frac {tensor\_size} {num\_devices} \times (num\_devices - 1) \times 4
  \label{dp-resync-cost}
\end{align}
\normalsize
%

This cost model assumes that the Allreduce operator is performed in the throughput-optimal ring topology~\cite{nccl, blink}.
In this approach, data is sent over the network twice, once to perform reduce scatter and then to perform all gather. 
The reduction operator for Allreduce is performed across the vector core/cores of each accelerator.
This is in line with a variety of prior work in the area of device placement~\cite{pip, piper, flexflow}.

%

\begin{figure}
\centering
\includegraphics[width=0.7\textwidth]{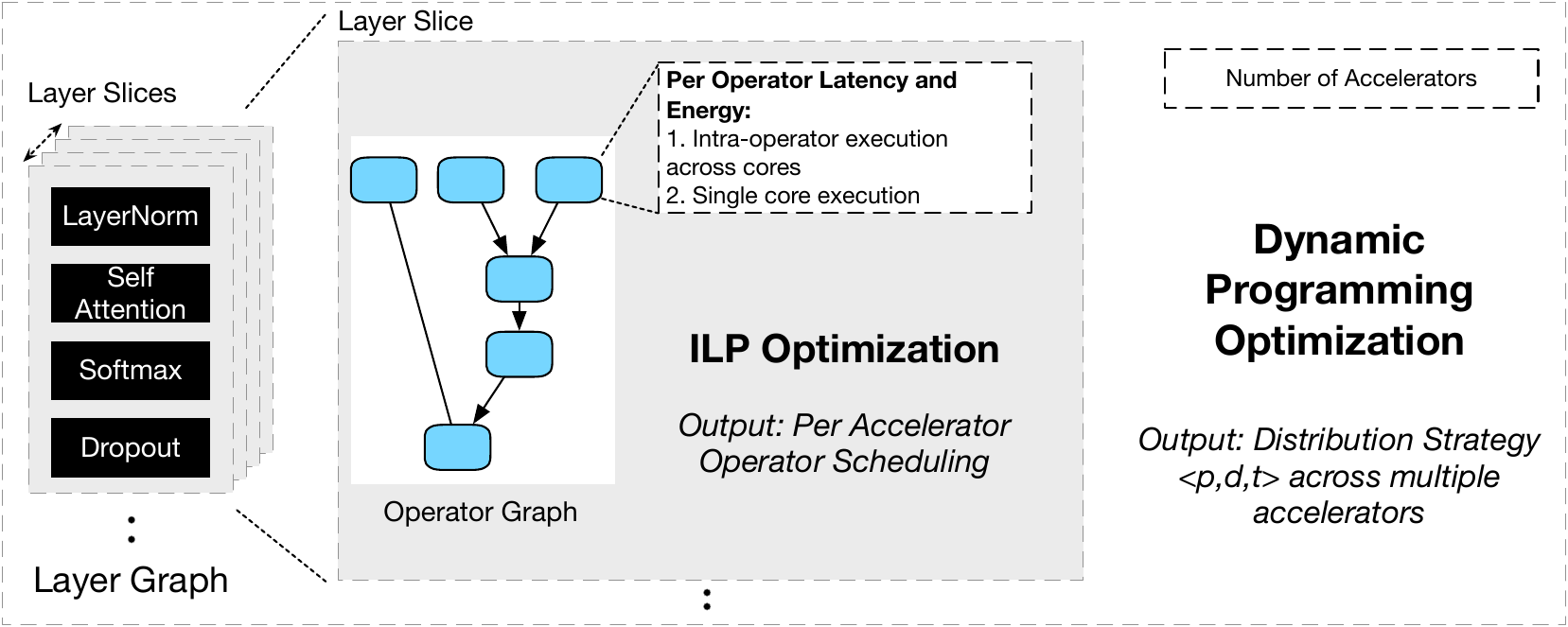}
\caption{The solver is executed for every architecture that is explored. It takes as input the layer graph and the corresponding operator graph. The ILP optimization solves to determine the optimal schedule and latency for every layer/layer slice. This information is used by the dynamic programming optimization to determine the training distribution strategy. }
\label{fig:algorithm}
\end{figure}

\subsection{Phaze Solver with ILP and Dynamic Programming Optimization}
\label{apdx:phaze_algorithm}

As models grow larger, determining accelerator architecture solely based on inference, as done in prior work, proves sub-optimal. This is due to the unique challenges presented by training, such as the fact that graphs are much larger than those used for inference, the optimizer and backward pass operators have distinct computational and memory requirements compared to forward pass operators, and training has a larger memory footprint.
It's important to note that the design of such accelerators depends not only on the model and its execution graph, but also the distribution strategy.

Traditionally, the device placement strategy used to distribute training execution relies on a fixed architecture to determine the execution time for each layer. This information is then used to establish the optimal number of stages in a pipeline, the layers that constitute each stage in a pipeline, data parallel width, and tensor model parallel width for end-to-end training.
However, this approach creates a cyclical dependency between device placement and architecture search optimization. On the other hand, the architecture search problem can be resolved if the distribution strategy is determined statically – either by manually identifying different modes and degrees of parallelism or using the memory footprint to balance execution. Nevertheless, this method is sub-optimal.
Thus, we devise an algorithmic solver that for every accelerator configuration determines the operator schedule and the model distribution strategy.

\phaze~employs an algorithmic solver for the following problem:
Given an accelerator configuration, what is the optimal operator schedule for the accelerator and the device placement strategy across multiple accelerators to distribute the training.
For the former, ILP solves the optimization problem without the resource constraint, ($z$) variables.
In case with the solution proposed, the resource is violated, then the ILP is resolved with all the constraints.

%
%
%
%

\section{Extended Evaluation and Ablation Studies}
\label{apdx:results}
\label{apdx:hysteresis}

\input{body/tables/workload_appendix}

\niparagraph{Model compute and memory properties.}
Table~\ref{tab:workloads_apdx} shows the evaluated models, the number of layers, parameter and activation size, and the compute complexity of the model.
Due to large activation size of OPT and GPT2 model, they require a larger High Bandwidth Memory (HBM) of 64GB to fit in the accelerator with activation stashing, or they require activation recomputation to run with a HBM size of 32GB. Meanwhile, Bertlarge and Bert with tensor model parallelism can execute at the optimal throughput with an HBM of just 32GB.
Tensor model parallelism allows another dimension of split that reduces the memory footprint. 
However, Megatron 8.3B and MegatronGPT3 still require activation recomputation to achieve the highest possible throughput.
For Megatron 2.5B, the maximum tensor model parallel width is 4, which is insufficient to store activations between the forward and backward pass on a device with only 32GB of HBM. Running with activation stashing becomes possible with an 80GB HBM, and achieves higher throughput than performing activation recomputation. 
In the case of MegatronGPT3, the model and activation sizes are so large that even with activation recomputation, it still requires a HBM of at least 64GB, and can only execute with activation recomputation, even with a 80GB HBM.
The OPT model exhibits a relatively high tensor and vector model complexity (given its size) due to the large sequence length.

\niparagraph{Reducing the overhead of executing the estimates.}
\phaze performs latency estimations for each combination of model and microbatch size. The estimations for each microbatch size are independent of one another, allowing them to be executed concurrently. This parallelization reduces the time required to run the estimator by a factor of four.
To further reduce the overhead of latency estimates, we leverage the repetitive structure of large language models and only estimate latencies for all the operators within a single layer or a layer slice.
All the non-repeat layers are estimated independently.
We also employ the same optimization for the ILP: it determines the optimal latency of a layer only once if it is repeated across the model.

\begin{figure}
\centering
\includegraphics[width=0.7\textwidth]{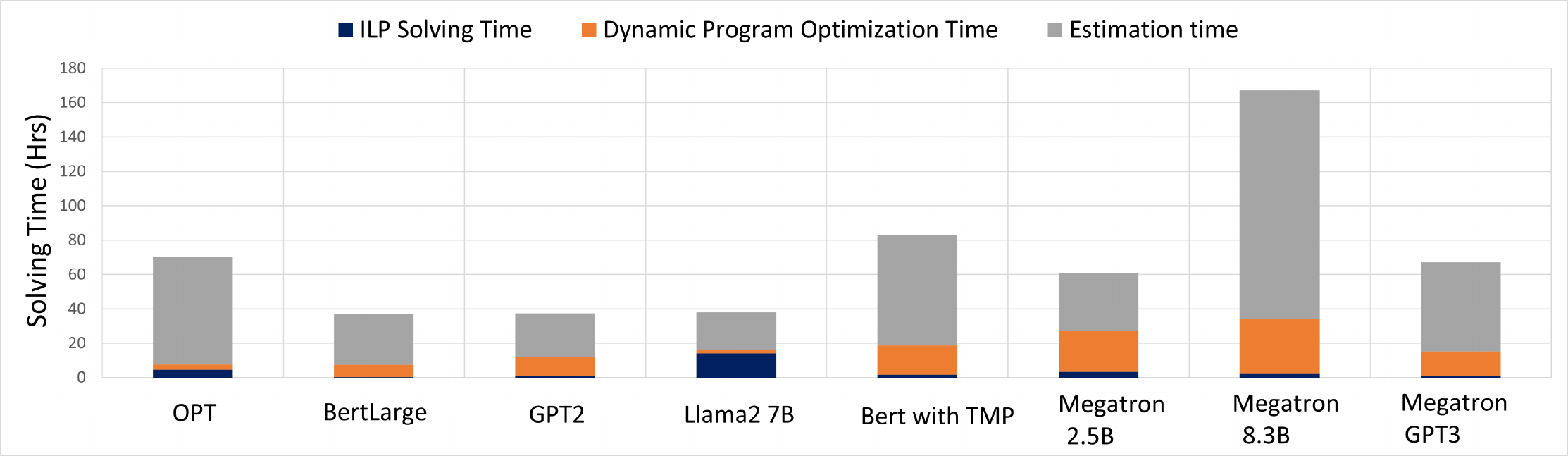}
\caption{The total execution time of each model and breakdown for ILP solving, Dynamic programming, and Estimation. }
\label{fig:execution_time}
\end{figure}

\begin{figure}
\centering
\includegraphics[width=0.7\textwidth]{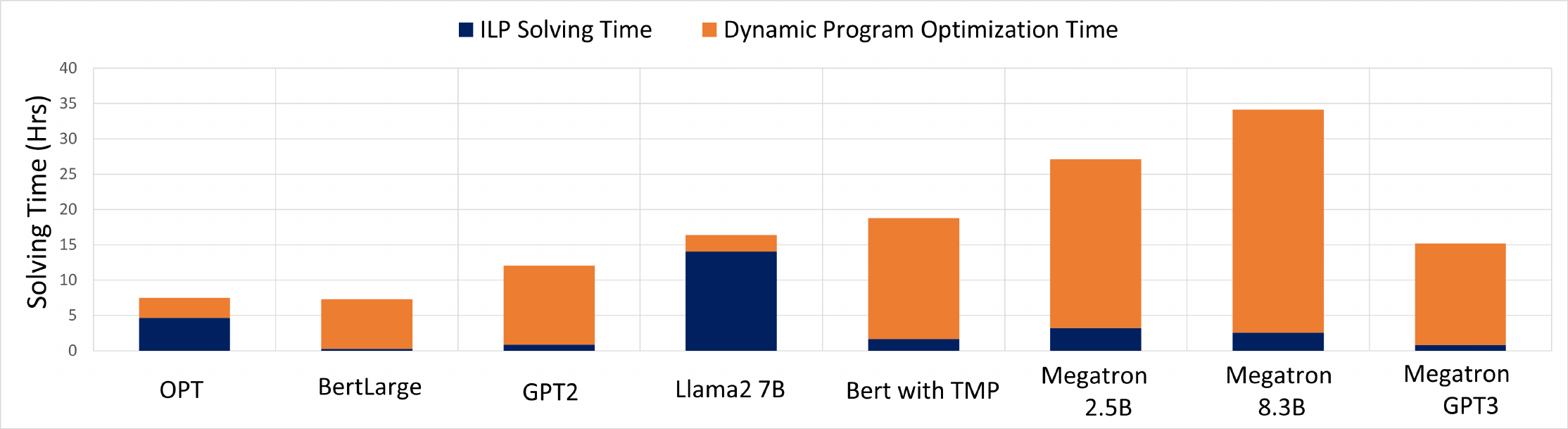}
\caption{Comparison of solving time split between ILP solving time and the dynamic program optimization time. ILP determines the optimal schedule per layer and layer slice, whereas the dynamic program determines the model partitioning.}
\label{fig:solving_time}
\vspace{-3ex}
\end{figure}

\niparagraph{Total optimization time.}
Figure~\ref{fig:execution_time} presents the total optimization time for each model, and the breakdown of the solver and estimation time.
This Figure has been presented with hysteresis 6.
The total convergence time, which includes estimation and solving time, is dependent on the hysteresis level, as this directly dictates the number of architectural configurations explored.
Overall, the estimation time increases proportionally with tensor sizes, whether for activation or parameters and the number of operators in the entire model graph.
The ILP solving time grows with number of operators in a layer graph and the number of tensor model widths.
A common property of large language models is replicated layers, and we make the ILP efficient as it reuses the optimal forward and backward latency from prior layer. 
The dynamic programming solving time is proportional to both the number of layers and the number of tensor model parallel widths.

The total convergence time is also dependent on the number of configurations of HBM and activation recomputation vs Stashing, that the model can run. Since MegatronGPT3 can only run with activation recomputation and HBM sizes of 64 and 80 GB, this significantly reduced the required total convergence time of the model, compared to Megatron8.3B. 
The exploration time for Megatron8.3B is higher than that for MegatronGPT3. This is because the former examines 1,2,4,8 tensor model parallel widths, unlike the latter which only explores 4 and 8. Additionally, MegatronGPT3 can only run with activation recomputation and HBM sizes of 64 and 80 GB, this significantly reduced the required total convergence time of the model, compared to other Megatron models.
Due to the size of the model, Llama2-7B could only run with microbatch sizes of 1 and 2, which significantly reduced the required total convergence time compared to other models.

\begin{figure}
\centering
\includegraphics[width=0.65\textwidth]{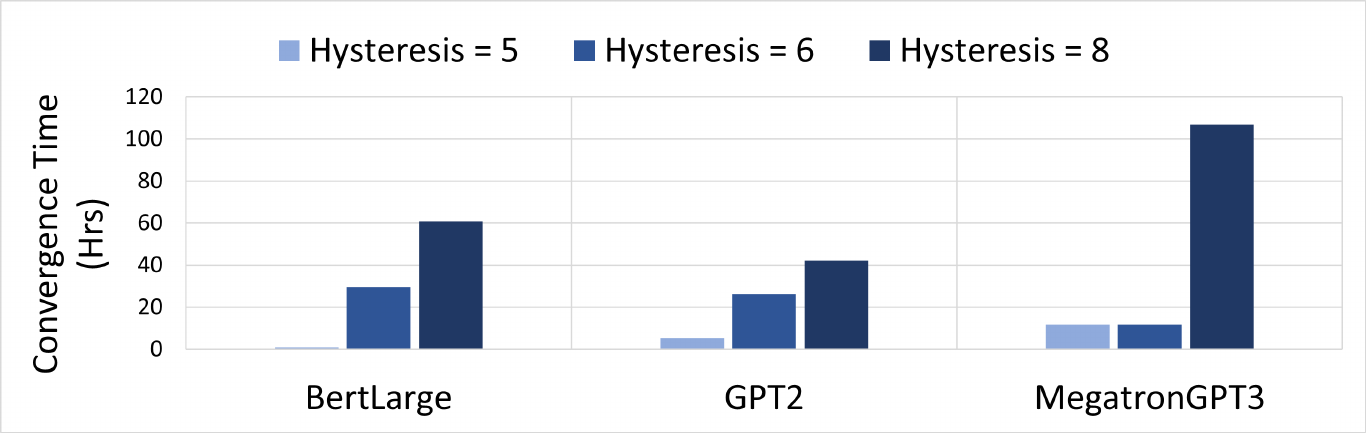}
\caption{ Convergence time for various hysteresis levels when running \phaze with just a single microbatch size and HBM configuration. Megatron models due to the added dimensionality of exploration (TMP) take longer to converge.}
\label{fig:convergence_time}
\end{figure}

\newcommand\filledcirc{\ensuremath{{\color{black}\bullet}\mathllap{\circ}}}

\begin{figure}[t!]
\centering
\includegraphics[width=0.5\textwidth]{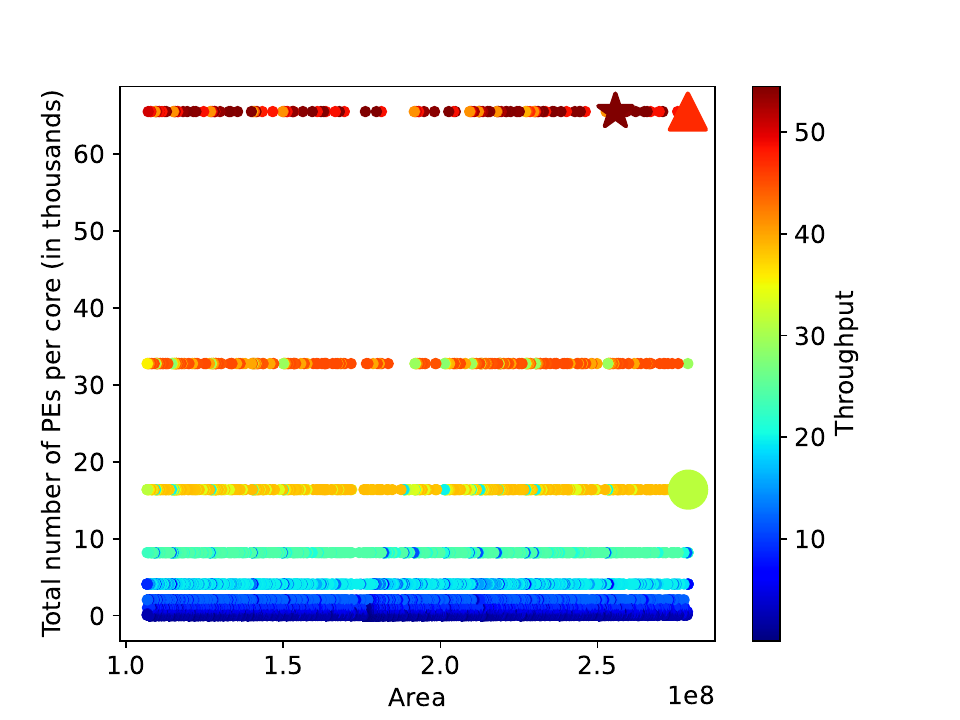}
\caption{The area, number of PEs per core, and throughput of each explored architecture by \phaze for BertLarge with microbatch size of 1. As the hysteresis level increases, the searched architecture's area decreases and eventually converges to an architecture with optimal throughput, \{2,512,256,256\}. The markers represent the architectures selected with each hysteresis level $\filledcirc$: H = 1,2 (selects the TPUv4 architecture), $\triangle$: H = 3, 4, $\star$: H = 6,7,8.}
\label{fig:bert_hysteresis}
\end{figure}

\niparagraph{Solving time breakdown.}
Figure~\ref{fig:solving_time} shows the execution time for the \phaze solver, divided into the ILP solving time and dynamic programming optimization.
While the dynamic programming optimization time predominates the solving times, it still takes significantly less percentage of the total convergence time than estimation. 
As mentioned earlier, the estimations for each microbatch size were conducted concurrently, leading to a substantial reduction in the required estimation time. In contrast, the solver navigates through each microbatch size sequentially. This sequential exploration contributes to an overall increase percentage in the convergence time spent running the solver. Furthermore, the dynamic programming optimization also explores various configurations for HBM and activation stashing vs. recomputation, leading to the dynamic programming optimizations occupying a larger portion of the overall execution time. For MegatronGPT3, the overall solving time is reduced, as it supports fewer HBM configurations and tensor model parallel widths compared to other models.

The ILP time is dependent on the number of tensor model widths supported as it needs to determine the optimal latency of layer slice per TMP width. 
However, the ILP time does not increase with the width of tensor model parallel model (4 vs 8), as the ILP only needs to establish a schedule for a single layer slice.
However, as models grow and the number of nodes increases, ILP's complexity generally rises.
The complexity of the models is outlined in Table~\ref{tab:workloads_apdx}.
The OPT model, despite its relatively small size, requires the longest solving time. This is dominated by the ILP due to violations of the $z$ constraints, and the ILP is recalculated for both the tensor and vector core resource constraints.

\niparagraph{Convergence time analysis and hysteresis study.}
Figure~\ref{fig:convergence_time} shows the convergence time with varying hysteresis levels when running \phaze with microbatch size 1 and an HBM of 80GB with Activation Recomputation.
As observed, the convergence time reduces by 11 $\times$ and 3 $\times$, respectively, when applying the heuristic described above with values of 5 and 6, in comparison to 8.
The exploration of Megatron models is more time-consuming due to the added dimensionality of tensor model parallelism, which necessitates latency estimations for all layer slices.
%
%
As previously stated, estimations on average account for up to 89\% of the convergence time. This motivates the pruning of accelerator configurations.

As Figure~\ref{fig:bert_hysteresis} shows, for hysteresis level 1 and 2, TPUv4 architecture is selected (as that is always explored because it has the largest area). 
With hysteresis level 3 and 4 we start to observe higher throughput architectures being selected, but that improvement diminishes when hysteresis is set 6, and does not improve further for hysteresis level 7 and 8. 
It's crucial to underscore that, even with the introduction of the heuristic, the accelerator configuration and the distribution strategy remain optimal at hysteresis values above 6. This is because, even when all feasible architectures are explored, the pruner consistently selects a design within 3\% of the largest areas amongst all feasible configurations. A hysteresis value of 6 or above consistently searches beyond the optimal architecture for each model.

\niparagraph{Sequential vs ILP operator scheduling.}
We compare the throughput of each model when utilizing \phaze's ILP solver to schedule operators versus employing a sequential schedule, both on the \phaze-Common architecture. 
Sequential scheduling executes every operator in intra-operator parallel. In contrast, our ILP scheduler performs an optimization across intra-operator scheduling and multi-operator scheduling.
We observe the most benefits using ILP scheduling for the Llama2 and BERT Large models with 50.6\% and 13.8\% throughput improvements respectively. 
OPT also observes a small throughput improvement compared to sequential scheduling.
For the remaining models, we observe the same throughput as with sequential scheduling.

There are two main reasons for this observation: 
First, many operators in these models are small enough that they do not fully utilize computational resources when running in intra-op parallel mode, therefore the latency of running intra-op is equivalent to running on a single core. Hence, these models benefit from running multiple operators in parallel or branching.
Second, the schedules for these models include multiple instances where operators with longer latencies, such as tensor core operators like matrix multiplication, can be scheduled to execute in parallel. In contrast, for models like GPT-2 and MegatronGPT-3, we do not observe instances where two tensor core operators are independent and can be executed in parallel. Only operators with orders of magnitude smaller latencies can be parallelized. Hence, the benefits of branching are overshadowed by the latency of the longer operators, leading to minimal improvements from the ILP scheduling.

\niparagraph{Model FLOPs utilization comparison.} 
In Table~\ref{tab:mfu} we present the Model FLOPS Utilization (MFU) of the TPUv4 (using expert Device Placement and using \phaze's solver) architecture and the \phaze common architecture. We follow the same formula to compute the MFU as presented in PaLM~\cite{palm} Appendix B.

When comparing architectures with the same effective tensor core FLOPs, higher observed throughput translates to higher MFU. As \phaze's common configuration has the same tensor core FLOPs as the TPUv4 configuration, its higher observed throughput translates to higher MFU. The key difference lies in the number of vector cores and global buffer memory size. \phaze prioritizes architectures with more vector cores within similar areas, albeit at the expense of reduced memory capacity. This choice enhances parallelization efficiency for operators like Layer Normalization.

However, it is important to note that \phaze searches across a variety of hardware architectures with different tensor core configurations, each with different peak matmul FLOPS. As MFU is the ratio of observed throughput to the theoretical maximum throughput of systems operating at peak matmul FLOPS, comparing systems with differing peak tensor core FLOPs means that a higher MFU does not always mean higher throughput, and vice versa.

\phaze currently optimizes for maximum throughput rather than utilization or MFU. Nonetheless, \phaze can be extended to enable users to optimize for other metrics, including throughput, utilization, or energy.

\begin{table*}[]
\centering
\footnotesize
\caption{Table details the Model FLOPs Utilization across TPUv4 architectures and the \phaze Common architecture. As Phaze's common configuration has the same tensor core FLOPs as the TPUv4 configuration, its higher observed throughput translates to higher MFU.}
\begin{tabular}{l|c|c|c}
\hline
\textbf{Model}         & \textbf{TPUv4 Expert DP (\%)} & \textbf{TPUv4 + Phaze Solver (\%)} & \textbf{Phaze Common (\%)} \\ \hline
OPT           & 5.4                           & 12.3                               & 17.1                       \\
BertLarge     & 15.9                          & 47.4                               & 88.1                       \\
GPT2          & 7.4                           & 19.6                               & 26.5                       \\
Llama2 7B          & 38.4                           & 39.8                               & 55.0                       \\
BERT with TMP & 13.7                          & 22.1                               & 57.4                       \\
Megatron 2.5B & 4.6                           & 4.9                                & 12.4                       \\
Megatron 8.3B & 6.4                           & 8.3                                & 17.2                       \\
MegatronGPT3  & 17.4                          & 17.5                               & 31.0                       \\ \hline
\end{tabular}
\label{tab:mfu}
\end{table*}

%% file: body/algorithms/phaze.tex
\begin{algorithm}
\caption{\phaze workflow algorithm}
\label{alg:phaze}

\newcommand{\funccommd}[1]{{\scriptsize\textcolor{DarkOrchid}{#1}}}
\newcommand{\mycommfont}[1]{{\scriptsize\itshape\textcolor{OliveGreen}{#1}}}

\scriptsize

\begin{algorithmic}
\STATE \textbf{Input} $models$  \quad \quad \quad \quad \quad \quad \quad \quad \quad \quad \quad \mycommfont{\# all models}
\STATE \textbf{Input} $num\_accelerator$ \quad \quad \quad \quad \quad \quad\mycommfont{\# maximum number of accelerators}
\STATE \textbf{Input} $area\_constraint$ \quad \quad \quad \quad \quad \quad\mycommfont{\# maximum area for each accelerator}
\STATE \textbf{Input} $architecture\_template$ \quad \quad \quad \mycommfont{\# search space for $cores_{tc},cores_{vc}, width, depth, glb, glb_{bw}, L2_{tc}, L2_{vc}$}

\STATE \textbf{Input} $training\_param$ \quad \quad \quad \quad \quad \quad \mycommfont{\# training parameters: $global\_batch\_size, mbs, hbm\_sizes$}

\STATE
\STATE \mycommfont{\# Generate all feasible architecture configurations}
\STATE $all\_valid\_acc\_configs$ = $\funccommd{get\_all\_configs\_within\_constraint}$($architecture\_template$, $area\_constraint$)
\STATE $sorted\_acc$ = $\funccommd{sort\_by\_area}(all\_valid\_acc\_configs)$

\STATE
\STATE \mycommfont{\# Generate graph $\forall$ $models$, $\forall$ TMP widths, $\forall$ micro-batch sizes}
\STATE $operator\_graphs$ = $\funccommd{extract\_graph}(model, training\_param.mbs\_list, training\_param.tmp\_list)$

\STATE
\STATE \mycommfont{\# Run solver to search for the best accelerator config $\forall$ models.}
\STATE $converged = False$
\STATE $config = sorted\_acc[0]$ \quad \mycommfont{\# start with largest config}
\WHILE {$not~converged$}
    \STATE $latency\_estimates\_ = \funccommd{get\_next\_estimate}(config, training\_param)$
        \STATE \mycommfont{\# Run ILP $\forall$ microbatch sizes.}
        \STATE $layer\_state = \funccommd{ilp}(operator\_graphs, latency\_estimates, training\_param, config)$
        \STATE \mycommfont{\# Run Dynamic Program $\forall$ microbatches, HBM sizes, and activation recomputation vs stashing.}
            \STATE $throughput, dp\_strategy = \funccommd{run\_dp}(layer\_state, config, is\_recompute, training\_param, num\_accelerator)$
    \STATE $\funccommd{append\_to\_explored\_acc\_list}(config, throughput, dp\_strategy)$
    \STATE $converged$, $config$ = $\funccommd{check\_converge}(hysteresis = 6, config)$
\ENDWHILE
\STATE
\STATE \mycommfont{\# Find the best configuration across $\forall$ models.}
\STATE $best\_acc, best\_dp = \funccommd{find\_highest\_tput\_acc}(models, explored\_acc\_list)$
\STATE \textbf{\textcolor{Blue}{return}} $best\_acc, best\_dp$

\STATE
\STATE \mycommfont{\# Helper function to check if search has converged}
\FUNCTION{$\funccommd{check\_converge}(hysteresis, curr\_config)$}    
        \STATE $curr\_avg\_t = \funccommd{avg\_across\_models}(throughput ~ in ~ explored\_acc\_configs)$
        \STATE $max\_thpt\_per\_area[curr\_config.area] = \funccommd{max}(
                max\_thpt\_per\_area[curr\_config.area], curr\_avg\_t)$
    
        \IF {$\funccommd{all\_configs\_in\_area\_explored}$($curr\_config.area$)}
        
            \STATE $next\_area = acc\_sorted[curr\_config.area\_idx + 1]$
    
            \STATE $larger\_area\_thgpt = [max\_thpt\_per\_area[area]$
                                     for area in $acc\_sorted$]
    
            \IF { $\funccommd{len}(larger\_area\_thgpt$)$ > hysteresis$}

                \STATE $larger\_area\_thgpt = larger\_area\_thgpt[-hysteresis:]$
                
                \STATE $converged = \funccommd{check\_if\_decreasing\_order}(larger\_area\_thgpt)$
    
               \STATE\textbf{\textcolor{Blue}{return}} $converged, next(config\_iter)$

            \ENDIF
        \ENDIF
        \STATE \textbf{\textcolor{Blue}{return}} $False, next(config\_iter)$
        
\ENDFUNCTION



\end{algorithmic}
\end{algorithm}

%% file: body/tables/op_mapping.tex
\begin{scriptsize}
\begin{table} [ht]
\centering
\caption{A sample of the operators, their mathematical equation, and core type mapping. One of the operands for Allreduce is obtained from the neighbouring device participating in the collective.}
\resizebox{0.8\columnwidth}{!}{
\scriptsize
\begin{tabular}{| l | l | l |}
 \hline
 \textbf{Operator} & \textbf{Mathematical Equation} & \textbf{Mapping} \\
 \hline
Convolution & $out(N_i,C_{out_j}) = bias(C_{out_j}) + \sum_{k=0}^{C_{in}-1} weight(C_{out_j},k) \cdot input(N_i,k)$ & Tensor Core \\
& & \\
  \hline 
 ReLU & $ ReLU(x) = max(0,x)$ & Vector Core \\
  & & \\
  \hline 
Linear & $ y = xA^{T} + b $ & Tensor Core\\
  & & \\
  \hline 
Softmax & $ Softmax(x_i) = \frac{exp(x_i)}{\sum_{j}{} exp(x_j)} $ & Vector Core \\
  & & \\
  \hline 
Batched MM & $ out = \beta X_1 + \alpha (X_2 \times W) $ & Tensor Core \\
  & & \\
  \hline
  Allreduce & $ out = Data1 + Data2 $ & Vector Core \\
  & & \\
  \hline
\end{tabular}}
\label{tab:mapping}
\end{table}
\end{scriptsize}

%% file: body/tables/workload_appendix.tex
\begin{scriptsize}
\newcommand\ExtraSep
{\dimexpr\cmidrulewidth+\aboverulesep+\belowrulesep\relax}

\newcolumntype{?}{!{\vrule width 2pt}}
\setlength\extrarowheight{3pt}

\begin{table}
\centering
\caption{Table details the complexity of the corresponding deep learning model, through number of operators per layer, number of layers, activation size, parameter size, Tensor Vector floating point operations (for TMP width 1, i.e., without tensor model parallelism), and Expert Device Placement Strategy expressed in the order of \{pipeline parallel depth (p), data parallel width (d), and TMP width (t)\}.}
\resizebox{1\columnwidth}{!}
{\begin{tabular}{ l | c | c | c | c | c | c | c}
 \hline 
\textbf{Model} & \textbf{\# of Layers} & \textbf{\# of Operators} & \textbf{Parameter Size} & \textbf{Activation Size} & \textbf{Tensor FLOPs} & \textbf{Vector FLOPs} & \textbf{Expert DP Strategy}\\
 \hline
 OPT~\cite{opt}       & 24 & 63 &  12 MB  & 4.7 GB & 2.5 $\times$ $10^{12}$ & 3.6 $\times$ $10^9$ &\{12, 85, 1\}
 \\
BertLarge~\cite{bert} & 24 & 34 & 23 MB & 609 MB & 21 $\times$ $10^9$  & 310 $\times$ $10^6$ & \{8, 128, 1\}
\\
GPT2~\cite{gpt}   & 48 & 101 & 152 MB & 5 GB & 557 $\times$ $10^9$ & 2.8 $\times$ $10^9$ & \{32, 32, 1\}
 \\
 Llama2 7B~\cite{llama2}   & 32 & 110  &  377 MB & 7.9 GB& 2.2 $\times$ $10^{12}$ & 9.9 $\times$ $10^9$& \{8, 128, 1\}
 \\
\hline
\multicolumn{6}{l}{\textbf{Tensor Model Parallel Models}} \\
\hline
Bert with TMP~\cite{bert} & 24 & 89 & 23 MB & 914 MB & 137 $\times$ $10^9$ & 301 $\times$ $10^6$ &  \{8, 128, 1\}
 \\
Megatron 2.5B~\cite{shoeybi2020megatronlm}   & 54 & 131 & 88 MB & 10 GB & 2.6 $\times$ $10^{12}$ & 2.6 $\times$ $10^9$ &  \{8, 32, 4\}
 \\
Megatron 8.3B~\cite{shoeybi2020megatronlm}   & 72 & 131 & 211 MB & 12 GB &  4.8 $\times$ $10^{12}$ & 4.8 $\times$ $10^9$ & \{8, 16, 8\}
 \\
MegatronGPT3~\cite{shoeybi2020megatronlm}      & 96 & 131 & 900 MB & 25 GB & 23 $\times$ $10^{12}$ & 9.6 $\times$ $10^9$ & \{32, 8, 4\}

\end{tabular}
\label{tab:workloads_apdx}
}

\end{table}
\end{scriptsize}